\documentclass[runningheads]{llncs}

\usepackage[mobile]{eccv}

\usepackage{graphicx}

\usepackage{booktabs}
\usepackage{tabularx}
\usepackage{arydshln}

\usepackage{siunitx}
\usepackage{etoolbox}
\robustify\bfseries

\usepackage{makecell}
\usepackage{adjustbox} %
\usepackage{array}
\usepackage{multirow}
\usepackage{rotating}
\usepackage[english]{babel}
\usepackage{pifont}
\usepackage[absolute]{textpos}

\usepackage[pagebackref=false,breaklinks=true,colorlinks,citecolor=eccvblue,bookmarks=true,bookmarksnumbered=true]{hyperref}
\hypersetup{
  pdftitle={Motion-prior Contrast Maximization for Dense Continuous-Time Motion Estimation},
  pdfsubject={Computer Vision, Robotics},
  pdfauthor={Friedhelm Hamann, Ziyun Wang, Ioannis Asmanis, Kenneth Chaney, Guillermo Gallego, Kostas Daniilidis},
  pdfkeywords={Event Cameras, Motion Estimation, Asynchronous Sensor, High Dynamic Range, High Temporal Resolution, Point Tracking}
}

\usepackage{orcidlink}

\definecolor{claude_red}{RGB}{255,0,0}

\newcommand{\unum}[2]{\multicolumn{1}{c}{\underline{\tablenum[table-format={#1}]{#2}}}}

\newcommand{\bnum}[1]{\bfseries #1}

\newcommand{\novalue}{{\textendash}}

\newcommand{\subsubsec}[1]{%
    \noindent\textbf{#1}%
}

\def\pol{p} %

\def\cE{\mathcal{E}} %
\def\numEvents{N_e} %
\def\numPixels{N_p} %
\def\numTraj{N_{\text{traj}}} %
\def\numCoeffs{N_c} %
\def\numBins{N_\text{bins}}
\def\numSubintervals{N_s}
\def\Warp{\mathbf{W}}

\def\bx{\mathbf{x}}
\def\bparams{\btheta} %

\def\tref{t_\text{ref}} %
\def\velflow{\mathbf{v}}

\def\bq{\mathbf{q}}

\def\cN{\mathcal{N}} %

\def\IWE{I}
\def\bx{\mathbf{x}}
\def\cE{\mathcal{E}}
\def\bparams{\boldsymbol{\theta}}
\def\balpha{\boldsymbol{\alpha}}
\def\Real{\mathbb{R}}

\def\Lnorm{L^1}

\newcommand{\cmark}{\ding{51}}%
\newcommand{\xmark}{\ding{55}}%

\usepackage{xcolor}
\definecolor{light-gray}{gray}{0.6}
\newcommand\gframe[1]{{\color{light-gray}\frame{#1}}}

\begin{document}

\title{Motion-prior Contrast Maximization for\\
Dense Continuous-Time Motion Estimation}

\titlerunning{Motion-prior Contrast Maximization}

\definecolor{somegray}{gray}{0.5}
\newcommand{\darkgrayed}[1]{\textcolor{somegray}{#1}}
\begin{textblock}{11}(2.5, -0.1)  %
\begin{center}
\darkgrayed{This paper has been accepted for publication at the
European Conference on Computer Vision (ECCV), 2024.
\copyright Springer
}
\end{center}
\end{textblock}

\author{
Friedhelm Hamann\inst{1}\orcidlink{0009-0004-8828-6919} \and
Ziyun Wang\inst{2}\orcidlink{0000-0002-9803-7949} \and
Ioannis Asmanis\inst{2}\orcidlink{0000-0001-8002-3197} \and
Kenneth Chaney\inst{2}\orcidlink{0000-0003-1768-6136} \and
Guillermo Gallego\inst{1,3}\orcidlink{0000-0002-2672-9241} \and
Kostas Daniilidis\inst{2,4}\orcidlink{0000-0003-0498-0758}
}

\authorrunning{F.~Hamann et al.}

\institute{TU Berlin and SCIoI Excellence Cluster, Berlin, Germany \and
University of Pennsylvania, Philadelphia, US \and
Einstein Center Digital Future and Robotics Institute Germany, Berlin, Germany \and
Archimedes, Athena RC, Greece
}

\maketitle

\begin{abstract}
Current optical flow and point-tracking methods rely heavily on synthetic datasets.
Event cameras are novel vision sensors with advantages in challenging visual conditions,
but state-of-the-art frame-based methods cannot be easily adapted to event data 
due to the limitations of current event simulators.
We introduce a novel self-supervised loss combining the Contrast Maximization framework with a non-linear motion prior in the form of pixel-level trajectories and propose an efficient solution to solve the high-dimensional assignment problem between non-linear trajectories and events.
Their effectiveness is demonstrated in two scenarios:
In dense continuous-time
motion estimation, our method improves the zero-shot performance of a synthetically trained model on the real-world dataset EVIMO2 by 29\%.
In optical flow estimation, our method elevates a simple UNet to achieve state-of-the-art performance among self-supervised methods on the DSEC optical flow benchmark.
Our code is available at \url{https://github.com/tub-rip/MotionPriorCMax}.
\end{abstract}

\section{Introduction}
\label{sec:intro}
\begin{figure*}[t]
	\centering  %
	\includegraphics[trim={1.5cm 0.8cm 1.3cm 0.9cm},clip, width=\linewidth]{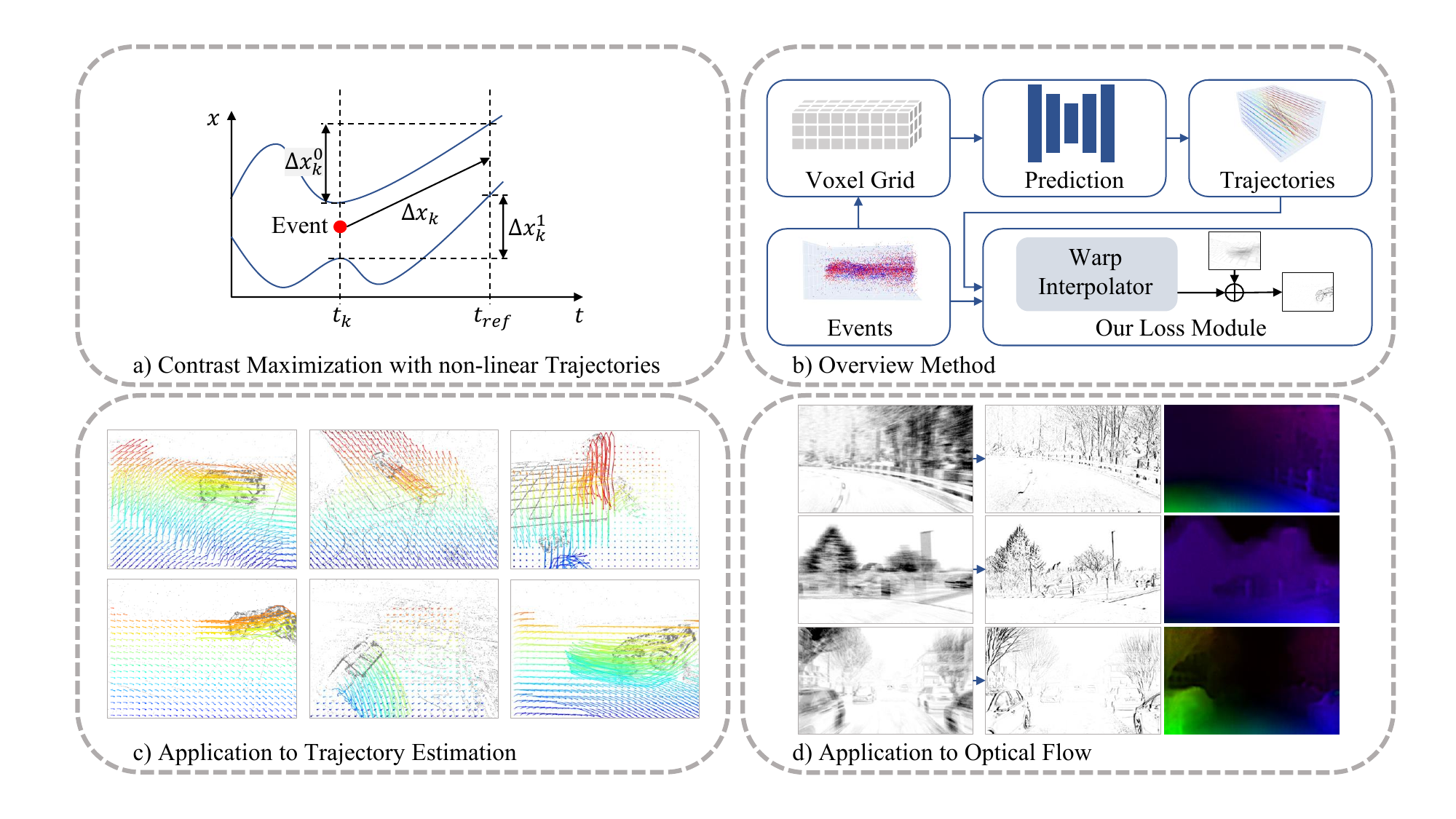}
	\caption{\emph{Summary}.
a) We present an approach to combine Contrast Maximization with dense non-linear trajectories.
b) We show how it can be used for self-supervised learning in a pipeline to predict dense point trajectories, 
and c) evaluate it on the EVIMO2 dataset, for which we generate dense point tracks.
d) Additionally, our approach provides state-of-the-art performance on self-supervised optical flow prediction.
}
\label{fig:eyecatcher}
\end{figure*}

Determining the motion of arbitrary projected world points on the image plane over long time intervals is a difficult low-level computer vision problem.
Researchers have studied it as optical flow and lately as point-tracking, with many practical applications in robotics, computational photography, video compression, and object-level tracking~\cite{Guney18eccvw}.
The best-performing frame-based methods for optical flow estimation and point tracking use large-scale synthetic datasets.
Synthetic data is diverse and has highly accurate ground truth (GT), but is unrealistic and methods trained on synthetic data show a sim-to-real gap.

Novel vision sensors called event cameras \cite{Lichtsteiner08ssc,Gallego20pami,Finateu20isscc} have emerged as promising alternatives to take on the problem.
Inspired by the transient visual pathway, which is responsible for motion perception, they are particularly fit for capturing scene motion
in the form of asynchronous pixel-wise intensity changes. 
This working principle endows them with advantages, such as high speed, high dynamic range (HDR) and low power consumption.

Event-based motion estimation methods can be categorized according to the complexity of the motion considered (i.e., number of degrees-of-freedom (DOF)) and to the solution strategy: model-based or learning-based. %
Low-DOF motions arise in sparse feature tracking and ego-motion estimation, while high-DOF motions describe more complex scenes, e.g., via per-pixel displacement (i.e., densely, over the whole image plane).
Focusing on the latter, the event-based optical flow problem has been extensively studied on mobile robotic datasets~\cite{Gehrig21ral,Zhu18ral}, where GT is calculated as the motion field (from known depth and poses).
Most approaches use this GT for direct supervision of dense flow prediction~\cite{Gehrig19iccv,Stoffregen20eccv,Gehrig21threedv,Wan22tip,Liu23iccv}.
Alternatively, several works use a contrast loss~\cite{Gallego19cvpr,Gallego18cvpr,Zhu19cvpr}, which allows training in a self-supervised manner.
However, the contrast loss is prone to undesired local optima, called event collapse \cite{Shiba22sensors,Shiba22aisy}, where many events are warped into a few pixels or lines.
Evaluation of event-based optical flow has been mostly done on the MVSEC \cite{Zhu18ral} and DSEC \cite{Gehrig21ral} datasets, which comprise largely uniform motions and test intervals from 0.022s to 0.1s. 
Recently, \cite{Gehrig24pami} has proposed a supervised method to predict pixel displacement over a larger duration (0.5s), by leveraging synthetic data. 
It relies on the generation of training event data from images, where current tools~\cite{Gehrig20cvpr} are not as mature as frame-based simulation, 
therefore it suffers from a sim-to-real performance gap.

In short, progress has been made and the research field is moving towards predicting motion over longer time intervals on the whole image plane, i.e., taking on more complex (i.e., nonlinear) motion problems.
This is precisely the problem tackled in this paper (\cref{fig:eyecatcher}): 
long-time and dense event-based motion estimation, reducing the domain adaptation gap of previous approaches.
It comes with several associated challenges, mainly overcoming the lack of large labeled datasets, 
and dealing with event noise and data association (events depend on motion, and for large motions, the appearance of ``corresponding'' events can be wildly different due to changes in motion direction, occlusions, etc.) while leveraging the space-time characteristics of event data.

To this end, we propose tackling the problem in two stages, by leveraging both supervised and self-supervised strategies: 
first, using supervised learning on synthetic data to provide initial model weights for motion estimation, 
and secondly, fine-tuning the network on real data via a self-supervised loss to reduce the domain adaptation gap.
Our technical contributions involve extending the contrast loss framework to regress continuous-time trajectories over long time intervals via motion priors (parametric functions that provide a good balance between motion generality and regularization \cite{Valmadre12cvpr,Wang21arxiv}).
This includes a solution to accurately and efficiently associate events to the trajectories (\cref{fig:method}).

More specifically, finding the association between events and motion trajectories (to warp corresponding events and achieve event alignment) is a high-dimensional problem (e.g., for a time window of 0.3 s one can consider about ten million events, and as many trajectories as pixels), which needs to be implemented in a differentiable and parallelizable manner.
We propose two actions to cope with these technical challenges.
First, we relax the problem by interpolating over a coarse spatio-temporal displacement field, which serves as a lookup table.
Secondly, we use a symbolic matrix framework~\cite{Feydy20neurips} to calculate the $K=\numTraj$ nearest neighbor (KNN) trajectories for each grid point, thereby solving KNN in a memory-efficient and differentiable way %
on GPU. 
The warp displacement at the grid point is then set to the average of the neighboring trajectories.

Our approach is versatile, allowing for different types of networks and trajectories.
Hence, we evaluate its performance on two applications:
dense continuous-time motion estimation and optical flow.
The results on EVIMO2 \cite{Burner22evimo2} show that fine-tuning with our self-supervised loss improves the zero-shot performance of a model pre-trained on synthetic data by 29\%.
On DSEC, our model shows state-of-the-art performance among self-supervised methods, on average improving the angular error by 19\% and the percentage of inliers by 14\%, while having a 5$\times$ faster inference time.
In summary, our contributions are (\cref{fig:eyecatcher}):
\begin{enumerate}

    \item We introduce motion priors (parametric functions with a good balance between generality and regularization) in the event-based contrast maximization framework for continuous-time and dense motion estimation.

    \item We combine the self-supervised loss with Bflow \cite{Gehrig24pami}, the current top-performing supervised model for dense continuous-time event-based motion estimation trained on synthetic data, and show that it can be used to improve over zero-shot performance on unseen real data (EVIMO2).

    \item We show that the combination of our loss with a simple U-Net architecture (à la EV-FlowNet \cite{Zhu18rss}) achieves state-of-the-art performance among the self-supervised methods on the DSEC Optical Flow benchmark. %
\end{enumerate}

\section{Related Work}
\label{sec:related}

\subsubsec{Frame-based Motion Estimation.}
Progress in deep learning triggered a large series of learned methods for optical flow estimation in classical, frame-based computer vision \cite{Dosovitskiy15iccv, Ilg17cvpr,Teed20eccv,Huang22eccv,Sun18cvpr,Bailer17cvpr}.
The solutions rely on large-scale synthetic datasets \cite{Mayer16cvpr,Dosovitskiy15iccv,Sun2021cvpr,Zen12eccv}.
Similarly related is the task of point tracking, which has shown impressive progress on frame-based data \cite{Sand08ijcv,Harley22eccv,Zheng23cvpr,Karaev23arxiv,Doersch23arxiv}, equally relying on simulated data \cite{Doersch22neurips,Zheng23cvpr,Butler12eccv}.
This approach is unsatisfying, as it is prone to out-of-distribution (OOD) problems that cannot be easily overcome.
As an alternative, self-supervised methods have been explored~\cite{Jonschkowski20eccv,Ren17aaai,Yu16eccvw,Janai18eccv} 
and shown to improve models pre-trained on simulated data for motion estimation tasks~\cite{Sun24arxiv,Stone21cvpr}.

\subsubsec{Event-based Optical Flow.}
Event cameras \cite{Lichtsteiner08ssc,Posch14ieee} are a relatively new technology, and have found applications in various computer vision domains, like mobile robotics~\cite{Weikersdorfer13icvs,Paredes24scirob,Guo24tro}, scene understanding~\cite{Hamann22icprvaib,Hamann24cvpr}, and computational imaging~\cite{Tulyakov22cvpr,Shiba23pami}.
Their exploration for low-level vision tasks has taken a similar trajectory in a compressed timeline as previously frame-based methods \cite{Gallego20pami}.
The first event-based optical flow methods were model-based~\cite{Benosman12nn,Benosman14tnnls,Orchard13biocas,Brosch15fns,Liu18bmvc,Shiba22eccv}, 
followed by learning-based approaches~\cite{Zhu18rss,Zhu19cvpr,Lee20eccv,Gehrig21threedv,Hagenaars21neurips,Ding22aaai,Paredes23iccv}.

The scarce availability of event data and less mature simulation technology compared to frame-based cameras are major obstacles~\cite{Gehrig20cvpr}.
Therefore, event-based optical flow has been mostly evaluated on data acquired through ego-motion, such as the robotics dataset MVSEC~\cite{Zhu18ral}, 
the driving dataset DSEC~\cite{Gehrig21ral} or the M3ED dataset~\cite{Chaney23cvprw}.
However, GT is calculated using the motion field equation (with data from a synchronized depth sensor) and therefore does not provide an accurate flow at the event rate, nor spatially at occlusions and independently moving objects (IMOs).
This GT is used to train supervised learning approaches \cite{Gehrig19iccv,Stoffregen20eccv,Gehrig21threedv,Wan22tip,Liu23iccv}, which inherit limitations from the ground truth.

Self-supervised methods for event-based optical flow lessen the dependency on GT labels by leveraging an event alignment error \cite{Zhu19cvpr,Ye19arxiv,Hagenaars21neurips,Shiba22eccv}.
Notably, variations of the contrast loss have been proposed~\cite{Gallego19cvpr}, however, so far the application has been limited to either low-DOF problems (e.g., feature tracking \cite{Gallego18cvpr,Seok20wacv,Chui21arxiv}, ego-motion \cite{Gallego17ral,Gallego18cvpr,Kim21ral}) or high-DOF but short-time optical flow problem \cite{Shiba22eccv,Paredes23iccv} (max. 0.1 s).
We expand the frontier to the challenging problem of long-time and high-DOF (complex) motion estimation, by exploiting trajectory priors.

\subsubsec{Trajectory Prior.} 
Using trajectories as motion priors has been a widely explored scheme in computer vision.
Specifically, \cite{Akhter10pami,Valmadre12cvpr,Zhu13pami} propose a linear combination of basis functions for structure from motion.
More recently it has been explored for dynamic novel view synthesis~\cite{Wang21arxiv}.
In the context of events, \cite{Tulyakov22cvpr} uses cubic motion splines for video frame interpolation, \cite{Gehrig24pami} uses B-splines for supervised learning of non-linear optical flow, \cite{Wang23arxiv_cc} use learned basis functions for event-based video decompression, and \cite{Seok20wacv,Chui21arxiv} show event-based feature tracking (low-DOF) using B\'ezier or B-spline curves in combination with a contrast loss.

\section{Methodology}
\label{sec:method}

\begin{figure*}[t]
	\centering  %
    \includegraphics[trim={0cm .6cm .15cm .6cm},clip, width=\linewidth]{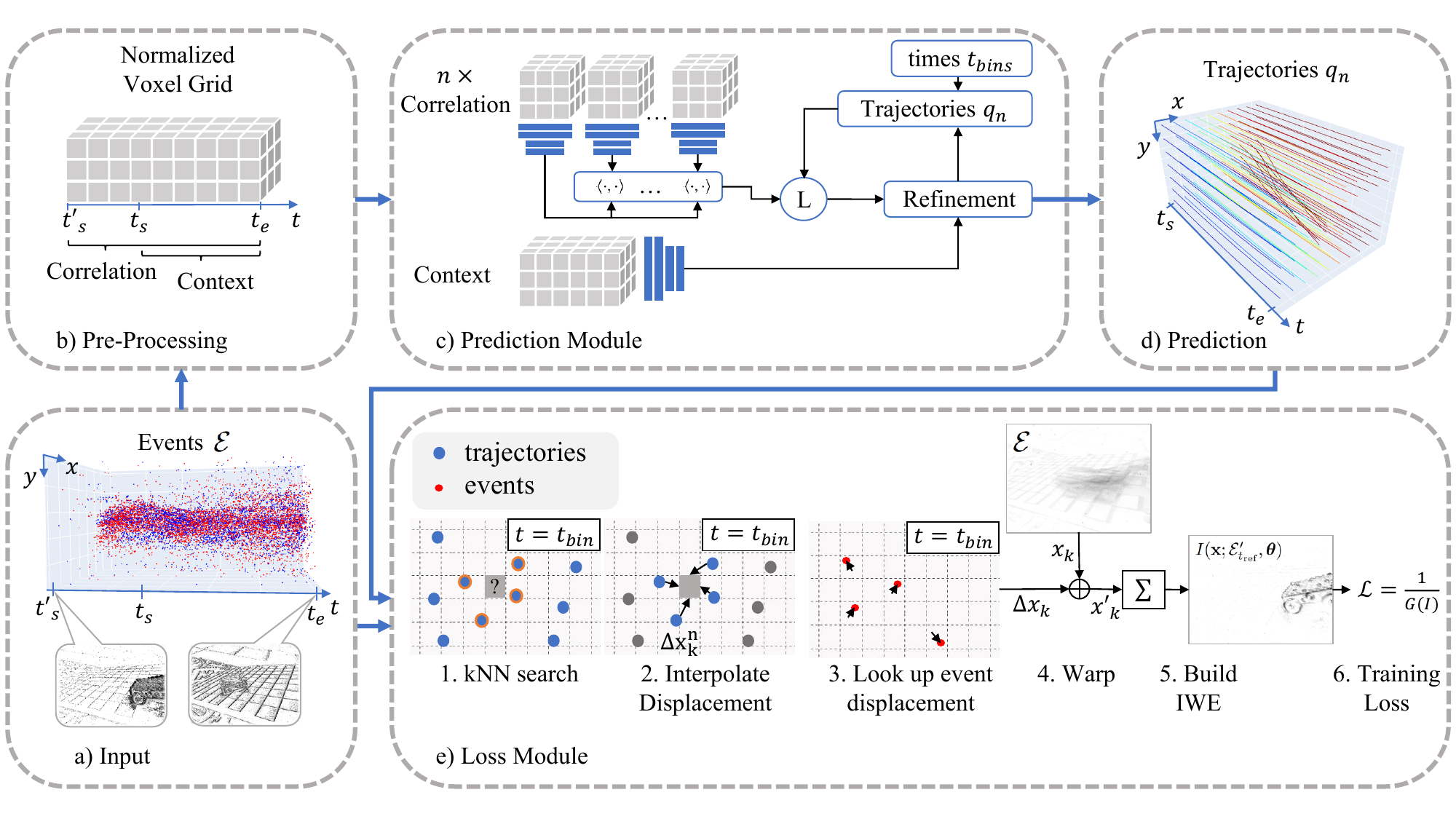}
	\caption{\emph{Pipeline overview}. 
 (a) Input events in a time interval 
 are (b) voxelized and (c) passed to an artificial neural network that predicts per-pixel coefficients for continuous-time trajectories (d).
 The raw events and predicted trajectories are fed to the loss module~(e).
 Here, a dense spatio-temporal displacement map is interpolated, and events are warped according to their looked-up displacement.
 Lastly, an image of warped events (IWE) is built at a random reference time and its gradient magnitude acts as training loss.
 Note that the prediction method displayed here is specific to the used Bflow backbone \cite{Gehrig24pami}, with additional events before the prediction start time $t_s$ as input.
 }
\label{fig:method}
\end{figure*}

\subsection{Contrast Loss}
Contrast Maximization (CM) \cite{Gallego18cvpr,Gallego19cvpr} is used for self-supervised training.
The idea is to find the point trajectories on the image plane best aligning with the events by maximizing the sharpness of warped events.
It is an iterative approach, with three steps per iteration:
($i$) Each event $e_k = (\bx_k, t_k, \pol_k)$ contains the pixel coordinates $\bx_k = (x_k,y_k)^\top$, timestamp $t_k$ and polarity $\pol_k$ of a brightness change of predefined size $C$ (contrast sensitivity). Events $\cE = \{ e_k \}_{k=1}^{\numEvents}$ are displaced according to a candidate motion hypothesis $\bx'_k = \Warp(\bx_k,t_k;\bparams)$, %
with parameters $\bparams$, to a reference time $\tref$, producing a set of warped events $\cE'_{\tref} = \{ e'_k \}_{k=1}^{\numEvents}$:
\begin{equation}
e_k \doteq (\bx_k,t_k,\pol_k) \;\,\stackrel{\Warp}{\mapsto}\;\,  e'_k \doteq (\bx'_k,\tref,\pol_k).
\end{equation}
Afterwards, ($ii$) the events are summed into an image of warped events (IWE),
\begin{equation}
\label{eq:IWE}
\textstyle
\IWE(\bx; \cE'_{\tref}, \bparams) \doteq 
\sum_{k=1}^{\numEvents} \cN (\bx;\bx'_k, \sigma^2), 
\end{equation}
which essentially counts the number of warped events $e'_k$ per pixel.
Lastly ($iii$) the contrast or sharpness of the IWE is computed (e.g., the gradient magnitude of the IWE,
$G \doteq \int \|\nabla I(\bx)\| d\bx$), which serves as a proxy for how well the model (motion hypothesis and $\bparams$) fit with the events produced by the true motion.

\subsection{Estimating Continuous-Time and Dense Motion Trajectories}
\newcommand{\warpOursOne}{$\bx'_k = \bx_k + \Delta \bx_k$}
\newcommand{\warpOursTwo}{$\Delta \bx_k = \frac{1}{\numTraj} \sum_n \Delta \bx_k^n$}

\newcommand{\warpSecretsOne}{$\bx'_k = \bx_k + \Delta \bx_k$}
\newcommand{\warpSecretsTwo}{$\Delta \bx_k = (\tref - t_k) \velflow(e_k)$}

\newcommand{\warpTamingOne}{$\bx'_k = \bx_k + \Delta \bx_k$}
\newcommand{\warpTamingTwo}{$\Delta \bx_k = \sum_i \bigl(\Delta t_{i} \velflow_i\bigr)(e_k)$}

\newcommand{\iweOurs}{$I(\bx) = \sum_k \cN(\bx; \bx'_k, \sigma^2)$}
\newcommand{\iweSecrets}{\iweOurs}
\newcommand{\iweTaming}{$T_{\pm}(\bx)= \frac{\sum_k t_k \cN(\bx; \bx'_k, \sigma^2)}{\sum_k \cN(\bx; \bx'_k, \sigma^2)}$}

\newcommand{\trefOurs}{$\sim \mathcal{U}(0, 1)$}
\newcommand{\trefSecrets}{$\{t_{0}, t_{0.5}, t_{1}\}$}
\newcommand{\trefTaming}{multi-partition $p$, multi-time-scale $s$}

\newcommand{\lossOurs}{$G(\tref)$}
\newcommand{\lossSecrets}{$\displaystyle \frac{G(t_0) + 2G(t_{0.5}) + G(t_1)}{4 G(\mathbf{v}=0)}$}
\newcommand{\lossTaming}{$\displaystyle \frac{1}{S}\sum_{s=0}^{S-1}\frac{1}{2^s}\sum_{p=0}^{2^s-1}\mathcal{L}_{\text{CM}, p}^{R/2^{s}}$}

\begin{table}[t]
\centering
\caption{Comparison of the contrast loss formulation in the most recent methods.}
\adjustbox{max width=\linewidth}{%
\setlength{\tabcolsep}{4pt}
\begin{tabular}{llll}
\toprule
Steps & Shiba et. al~\cite{Shiba22eccv} & Paredes et. al~\cite{Paredes23iccv} & Ours\\ 
\midrule 
Warp displacement \eqref{eq:warpOFlow}: &  \warpSecretsTwo & \warpTamingTwo & \warpOursTwo \\[5pt]
               &  linear trajectory & flow concatenation & \textbf{non-linear trajectory} \\[5pt]
IWE            &  \iweSecrets     & \iweTaming & \iweOurs \\[8pt]
Reference time $\tref$:  &  \trefSecrets    & \trefTaming   & \trefOurs \\[5pt]
Contrast objective  &  \lossSecrets    & \lossTaming  & \lossOurs \\
\bottomrule
\end{tabular}
}
\label{tab:method:loss_overview}
\end{table}
The CM framework has been extended in several works.
\Cref{tab:method:loss_overview} compares recent extensions for optical flow estimation, 
where events are warped as
\begin{equation}
\label{eq:warpOFlow}
\bx'_k = \bx_k + \Delta \bx_k.
\end{equation}
The main differences between these approaches lie in the event displacement (i.e., warp) model, the reference times used, and the loss function (i.e., event-alignment metric), with the design choices having two goals: fitting the event data and regularizing the solution (e.g., avoiding event collapse \cite{Shiba22sensors}).
Previous methods used mostly a linear model \cite{Shiba22eccv,Shiba24pami} or partitioned the inference interval into smaller flow intervals so that its concatenation need not be linear \cite{Paredes23iccv}.
By contrast, we introduce an explicit continuous-time non-linear trajectory model, a random reference time per iteration, and an easier contrast loss.

\subsubsec{Trajectory Representation.}
\Cref{fig:method} shows an overview of the training pipeline.
Events in the time interval $[t_s, t_e]$ are %
fed into a neural network that makes per-pixel predictions of the parameters (coefficients) $\bparams \equiv \balpha_n=(\alpha_{n,1},\ldots,\alpha_{n,\numCoeffs})^\top \in \Real^{\numCoeffs}$ of a continuous-time trajectory 
$\bq_n(t; \balpha_n) \equiv \bq_n(t) = (x_n(t), y_n(t))^\top$.
There is one trajectory per pixel (i.e., ``dense'' character), $n = 1, \ldots, \numPixels$, where $\numPixels = hw$ is the number of pixels (image height $h$ and width $w$). 
Thus, $n$ is the spatial index of the trajectory, identifying it among all trajectories on the image plane. 

We model trajectories as weighted combinations of basis functions, $\bq_n(t) = \sum^{\numCoeffs}_{j=1} g_j(t) \mathbf{p}_{n,j}$, 
where $g_j(t)$ are temporal basis shared by all trajectories, 
and $\mathbf{p}_{n,j} = (\alpha^x_{n,j}, \alpha^y_{n,j})^{\top}$ are ``control points'' 
(we write $x$ and $y$ components explicitly with separate coefficients in $\balpha_n$).
We investigate polynomial basis $g_j(t) = t^j$,
B\'ezier curves
with basis $g_j(t) = \binom{\numCoeffs}{j}(1-t)^{\numCoeffs-j}t^j$,
as well as a learned basis.

\subsubsec{Spatio-Temporal Event Warping.}
The trajectories can be used to warp events. 
This would require finding the trajectory that passes through the space-time coordinates of the event and then finding the value of the trajectory at the reference time (i.e., the warped event location). 
However, the association between events and trajectories is unknown and needs to be estimated simultaneously with the parameters (i.e., shape) of the trajectories. 
We circumvent the problem by using a soft association between events and trajectories.
Each event $e_k$ is associated with its $\numTraj$ nearest neighboring trajectories $\{\bq_n\}_{n=1}^{\numTraj}$, 
and the event displacement $\Delta\bx_k$ in \eqref{eq:warpOFlow} is computed as the average of the respective trajectory displacements $\{\Delta \bx_k^n\}_{n=1}^{\numTraj}$.
The trajectory displacement $\Delta \bx_k^n$ is defined as the difference of trajectory locations at the time of the event and the reference time. 
The event displacement is defined as the mean:
\begin{equation}
    \label{eq:eventdisplacement}
    \Delta \bx_k \doteq \frac{1}{\numTraj} \sum_{n=1}^{\numTraj} \Delta \bx_k^n, 
    \quad \text{with}\quad \Delta \bx_k^n \doteq \bq_n(\tref) - \bq_n(t_k).
\end{equation}

\subsubsec{Loss Calculation.}
Once the events are warped using \eqref{eq:warpOFlow}, it is straightforward to compute the IWE.
We adopt the magnitude of the IWE gradient as loss function \cite{Gallego19cvpr,Shiba24pami}.
Moreover, we use the contrast loss in a self-supervised learning setting, choosing a different reference time $\tref$, uniformly sampled in the observation interval, for every batch during training. 
This simplifies the objective to a single warping operation and loss calculation (e.g., compared to the three warping operations and loss calculations in the optimization-based approach by \cite{Shiba22eccv,Shiba24pami}), while it has added regularization benefits, as mentioned below.

\subsubsec{Memory Effective Computation of the Displacement Field.}
The number of events and trajectories can be very large, and even more their combination.
Calculating KNN for every event can quickly become computationally unfeasible, and traditional algorithms cannot be efficiently implemented in deep-learning frameworks.
We propose calculating the per-event displacement $\Delta \bx_k$ in \eqref{eq:warpOFlow} by first interpolating a dense but coarser spatio-temporal displacement field, and then looking up the per-event displacement in such space-time volume.
Moreover, we relax the problem by solving the KNN search in 2D instead of in the volume. %

\Cref{fig:method} shows an overview of the interpolation process.
The displacement map is a tensor of shape $[\numBins, h/4, w/4]$, where $\numBins$ is the number of temporal bins, and $h$ and $w$ are the sensor's height and width, respectively.
For each temporal bin-center $t_\text{bins}$ the trajectories $\bq_n(t=t_\text{bins})$ are calculated and the KNN interpolation is performed between each pixel of a channel $\bx[t_\text{bins}]$ and $\bq_n(t=t_\text{bins})$.
We implement the KNN approach using KeOps~\cite{Feydy20neurips}, a framework for symbolic matrix computation, which provides a memory-efficient and differentiable solution.
We perform a KNN search for every voxel in the table, and events are warped after a lookup operation.
Please note that this step is entirely independent of the voxel grid passed as input to the backbone, and all raw events are used to calculate the contrast loss.
From the calculated displacement map, we can directly look up per-event displacements $\Delta \bx_k$.

\subsubsec{Regularization.}
Objectives based on event alignment are prone to undesired local minima called ``event collapse'' \cite{Shiba22sensors,Shiba22aisy}. %
Our formulation inherently provides regularization, temporally by the smoothness of the motion prior, and spatially by interpolating the event flow from several trajectories via a soft assignment.
Our framework uses two additional regularization sources.
Firstly, we penalize the magnitude of the spatial gradient of the interpolated displacement field between consecutive timesteps, $R \doteq \| \nabla(\Delta \bq_*(t))) \|_{L^1(\Omega)}$. 
To interpolate $\bq_*$ from the sparse trajectories $\bq_n$, we use the same KNN indices as for the main volume.
This loss encourages spatial smoothness of the trajectories.

Secondly, we use a multi-reference formulation by randomly sampling a reference time at every training step, while previous work used multiple and fixed timestamps (\cref{tab:method:loss_overview}).
This has the advantage that ($i$) IWEs are required to be sharp at truly \emph{any} time 
and ($ii$)
the memory requirement is lowered as it scales linearly with the number of reference times used during a training step.

In summary, the training loss is 
\begin{equation}
\mathcal{L} = 1/G + \lambda\,R,
\end{equation}
with $\lambda>0$ as regularization weight. 
Further details are in the supplementary.

\subsubsec{Pre-processing and Prediction-Module Architecture.}
Events record brightness changes asynchronously, in the form of a sparse spatio-temporal signal.
As input to the Prediction Module, events are customarily converted to voxel grids~\cite{Zhu19cvpr} for compatibility with conventional artificial neural networks.
An event volume is discretized in the time dimension, and each voxel counts the number of events within it (bi-linearly voted for undistorted, rectified events).

In principle, our self-supervised loss module can be paired with any segmentation or optical flow network architecture.
We use a U-Net architecture for the experiments on DSEC (\cref{sec:exp:oflow}), and the architecture in the recent Bflow method~\cite{Gehrig24pami}, inspired by RAFT~\cite{Teed20eccv}, for the experiments on EVIMO2 (\cref{sec:exp:traj_results}).

\section{Experiments}
\label{sec:experiments}

We test our method on two applications.
First, we test the capabilities for non-linear trajectory estimation
on the real-world dataset EVIMO2~\cite{Burner22evimo2}, with additional results on the synthetic MultiFlow dataset~\cite{Gehrig24pami}, which we use for pre-training (\cref{sec:exp:traj_results}).
Secondly, we evaluate our method on the DSEC dataset~\cite{Gehrig21ral,Gehrig21threedv} 
because this has been the previous frontier for event-based optical flow estimation and it allows for direct comparison with prior work (\cref{sec:exp:oflow}). 
Additional results, including on the MVSEC dataset \cite{Zhu18ral}, are given in the supplement.

\subsection{Results on Non-Linear Trajectory Estimation}
\label{sec:exp:traj_results}

\begin{table}[t]
\centering
\caption{Results on EVIMO2 \cite{Burner22evimo2}.}
\label{tab:exp:results_evimo}
\adjustbox{max width=0.75\linewidth}{
\setlength{\tabcolsep}{6pt}
\begin{tabular}{ll*{3}{S[table-format=2.3]}}
\toprule
 & \textbf{Method} & {\textbf{TEPE} $\downarrow$} & {\textbf{TAE} $\downarrow$} 
 & {\textbf{\%Out} $\downarrow$}
 \\
\midrule

& {Paredes et. al~\cite{Paredes23iccv}}
 & 21.6887 & 51.9079 & 0.634 \\

& {E-RAFT (linear)~\cite{Gehrig21threedv}}
 & 19.384410858154297 & 74.51759338378906 & 0.6563485264778137 \\

& {BFlow (zero-shot)~\cite{Gehrig24pami}}
 & 8.62916088104248 & 19.9359054565429 & 0.362601906061172 \\

& {BFlow (in-domain)}
 & \bnum{3.38006925582885} & \bnum{11.6800241470336} & \bnum{0.166480526328086} \\

& {Ours (self-supervised)}
 & \unum{1.2}{6.14132022857666} & \unum{2.2}{16.9782161712646} & \unum{1.2}{0.254222095012664} \\

\bottomrule
\end{tabular}
}
\end{table}

\def\figWidth{0.185\linewidth}
\begin{figure*}[ht!]
	\centering
    {\scriptsize
    \setlength{\tabcolsep}{2pt}
	\begin{tabular}{
	>{\centering\arraybackslash}m{0.3cm} 
	>{\centering\arraybackslash}m{\figWidth} 
	>{\centering\arraybackslash}m{\figWidth} 
	>{\centering\arraybackslash}m{\figWidth} 
	>{\centering\arraybackslash}m{\figWidth} 
    }

        \rotatebox{90}{\makecell{Example 1}}
		&\gframe{\includegraphics[clip,trim={0cm 0cm 0cm 0cm},width=\linewidth]{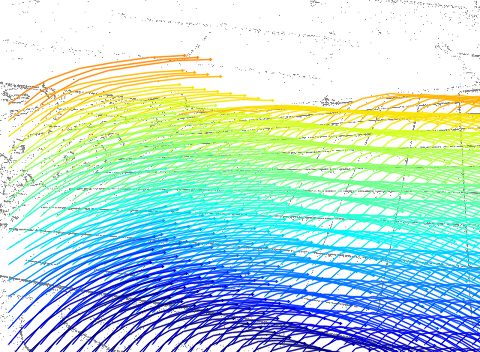}}
        &\gframe{\includegraphics[clip,trim={0cm 0cm 0cm 0cm},width=\linewidth]{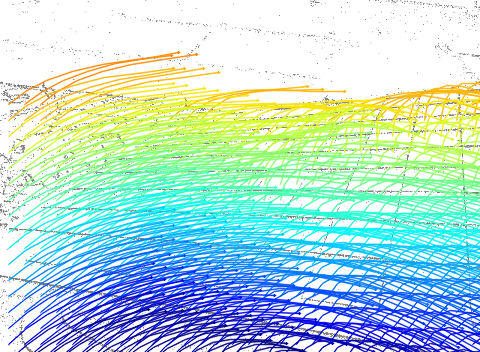}}
		&\gframe{\includegraphics[clip,trim={0cm 0cm 0cm 0cm},width=\linewidth]{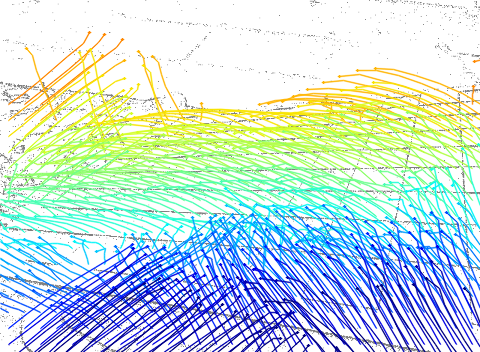}}
        &\gframe{\includegraphics[clip,trim={0cm 0cm 0cm 0cm},width=\linewidth]{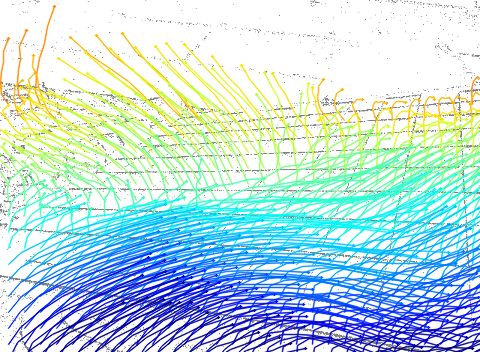}} \\

        \rotatebox{90}{\makecell{Example 2}}
		&\gframe{\includegraphics[clip,trim={0cm 0cm 0cm 0cm},width=\linewidth]{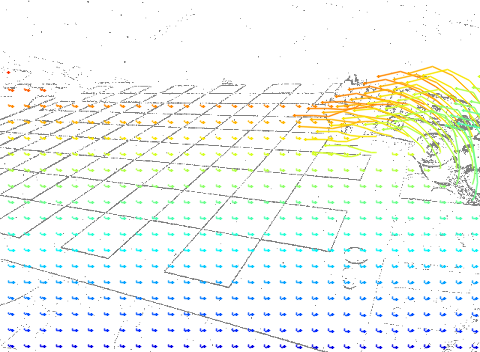}}
        &\gframe{\includegraphics[clip,trim={0cm 0cm 0cm 0cm},width=\linewidth]{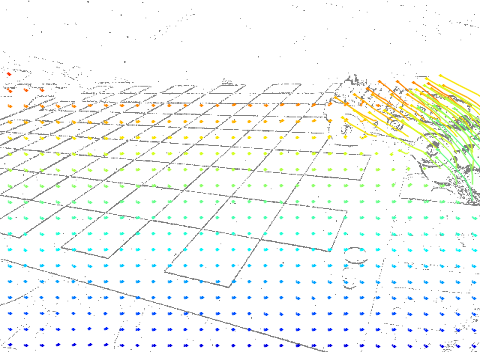}}
		&\gframe{\includegraphics[clip,trim={0cm 0cm 0cm 0cm},width=\linewidth]{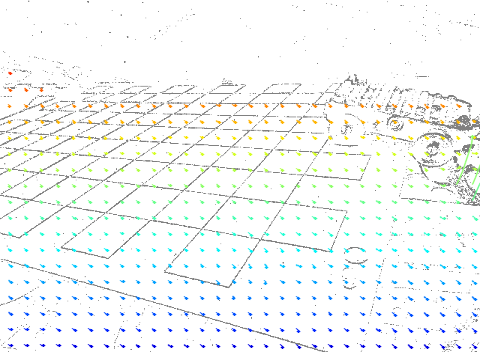}}
        &\gframe{\includegraphics[clip,trim={0cm 0cm 0cm 0cm},width=\linewidth]{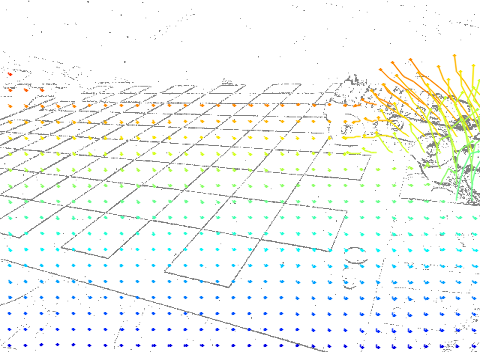}} \\

        \rotatebox{90}{\makecell{Example 3}}
		&\gframe{\includegraphics[clip,trim={0cm 0cm 0cm 0cm},width=\linewidth]{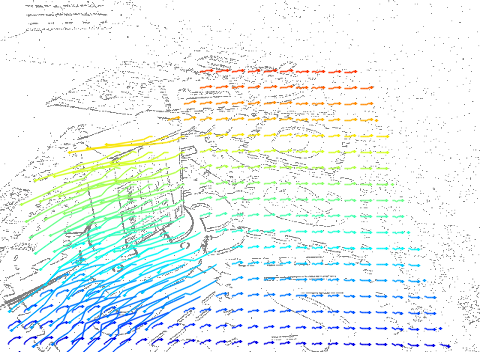}}
		&\gframe{\includegraphics[clip,trim={0cm 0cm 0cm 0cm},width=\linewidth]{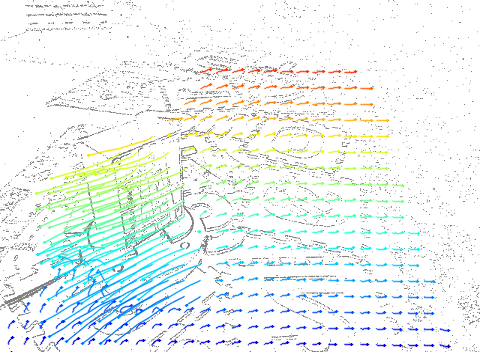}}
		&\gframe{\includegraphics[clip,trim={0cm 0cm 0cm 0cm},width=\linewidth]{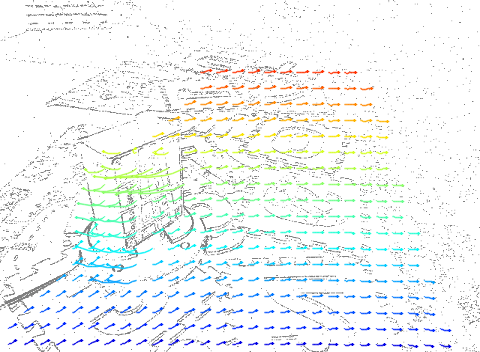}}
        &\gframe{\includegraphics[clip,trim={0cm 0cm 0cm 0cm},width=\linewidth]{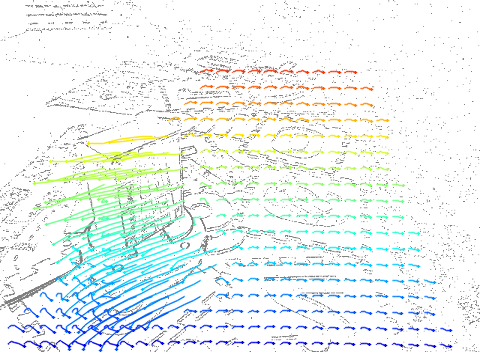}} \\

        \rotatebox{90}{\makecell{Example 4}}
		&\gframe{\includegraphics[clip,trim={0cm 0cm 0cm 0cm},width=\linewidth]{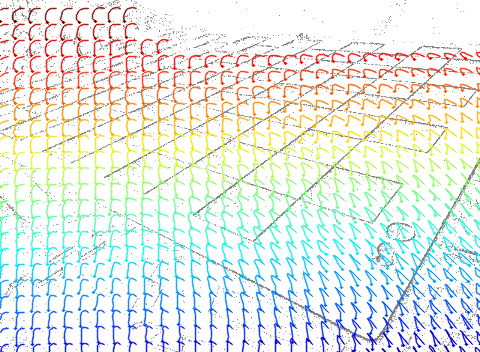}}
		&\gframe{\includegraphics[clip,trim={0cm 0cm 0cm 0cm},width=\linewidth]{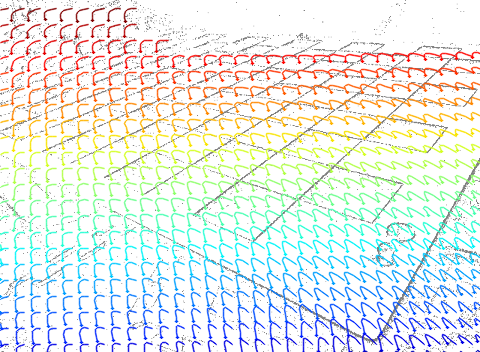}}
		&\gframe{\includegraphics[clip,trim={0cm 0cm 0cm 0cm},width=\linewidth]{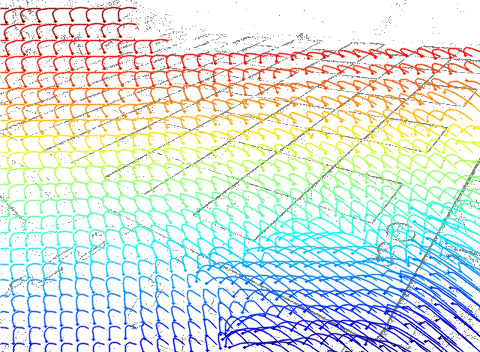}}
        &\gframe{\includegraphics[clip,trim={0cm 0cm 0cm 0cm},width=\linewidth]{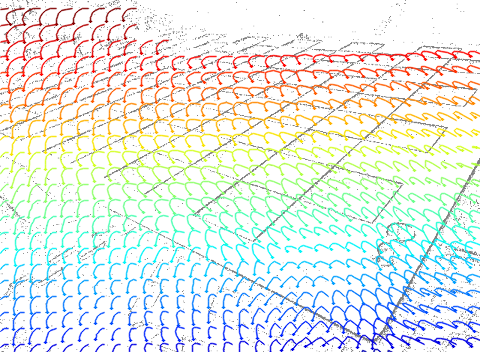}} \\

        \rotatebox{90}{\makecell{Example 5}}
		&\gframe{\includegraphics[clip,trim={0cm 0cm 0cm 0cm},width=\linewidth]{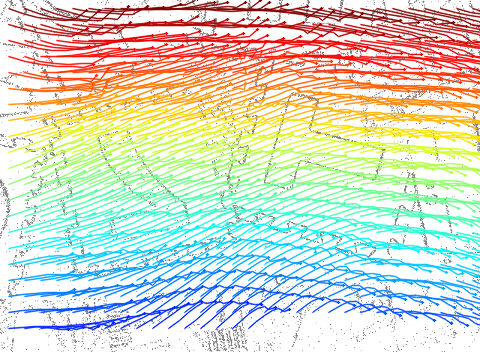}}
		&\gframe{\includegraphics[clip,trim={0cm 0cm 0cm 0cm},width=\linewidth]{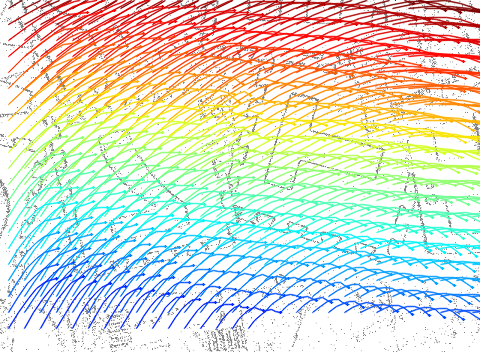}}
		&\gframe{\includegraphics[clip,trim={0cm 0cm 0cm 0cm},width=\linewidth]{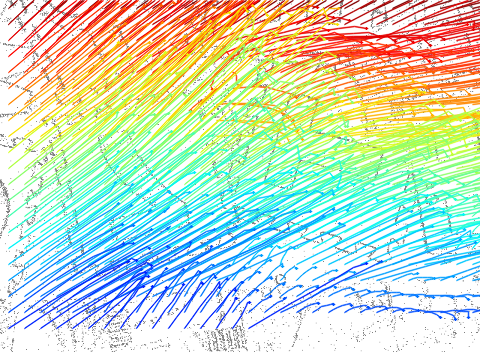}}
        &\gframe{\includegraphics[clip,trim={0cm 0cm 0cm 0cm},width=\linewidth]{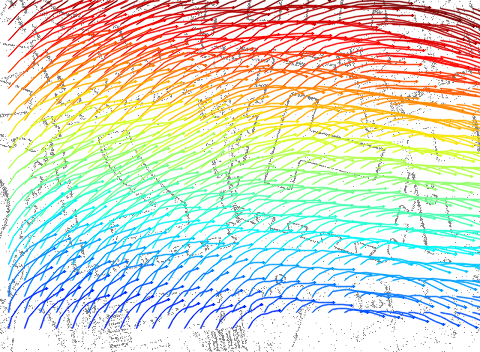}} \\

        \rotatebox{90}{\makecell{Example 6}}
		&\gframe{\includegraphics[clip,trim={0cm 0cm 0cm 0cm},width=\linewidth]{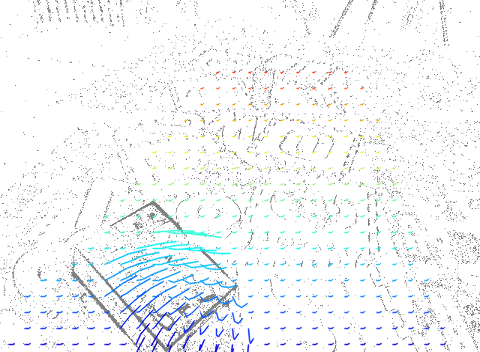}}
		&\gframe{\includegraphics[clip,trim={0cm 0cm 0cm 0cm},width=\linewidth]{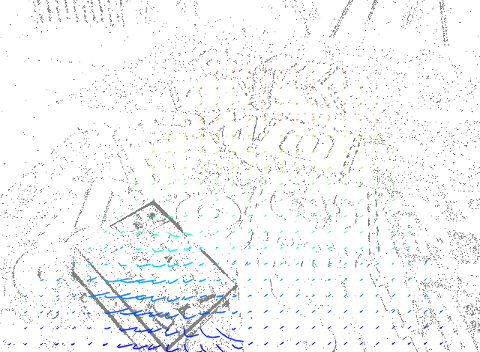}}
		&\gframe{\includegraphics[clip,trim={0cm 0cm 0cm 0cm},width=\linewidth]{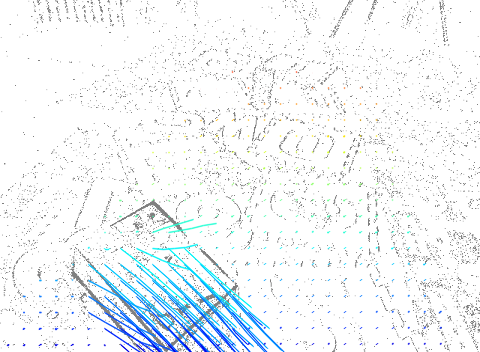}}
        &\gframe{\includegraphics[clip,trim={0cm 0cm 0cm 0cm},width=\linewidth]{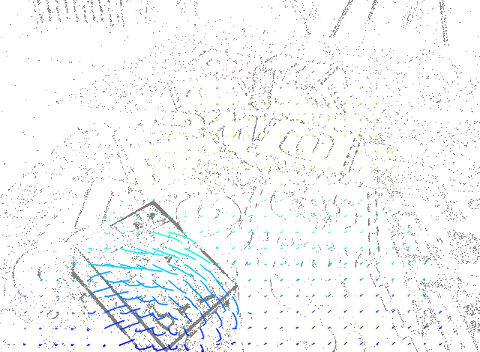}} \\

		& \textbf{(a)} GT
		& \textbf{(b)} In-domain
		& \textbf{(c)} Zero-shot
		& \textbf{(d)} Ours
	\end{tabular}
	}
	\caption{Visualization of predicted trajectories on EVIMO2 data. 
 \emph{GT}: Ground truth.
 \emph{In-domain}: fine-tuned on EVIMO2 using GT (supervised). 
 \emph{Zero-shot}: network trained only on synthetic data (out-of-domain prediction). 
 \emph{Ours}: Pre-trained on synthetic data, fine-tuned with self-supervised loss. 
 Note that supervision in-domain is often impossible in practice because dense trajectory labels for real data are difficult to obtain.}
	\label{fig:exp:results_evimo2}
\end{figure*}

\begin{figure*}[t]
	\centering  %
    \includegraphics[trim={0.5cm 0cm 2.75cm 2.5cm},clip, width=0.42\linewidth]{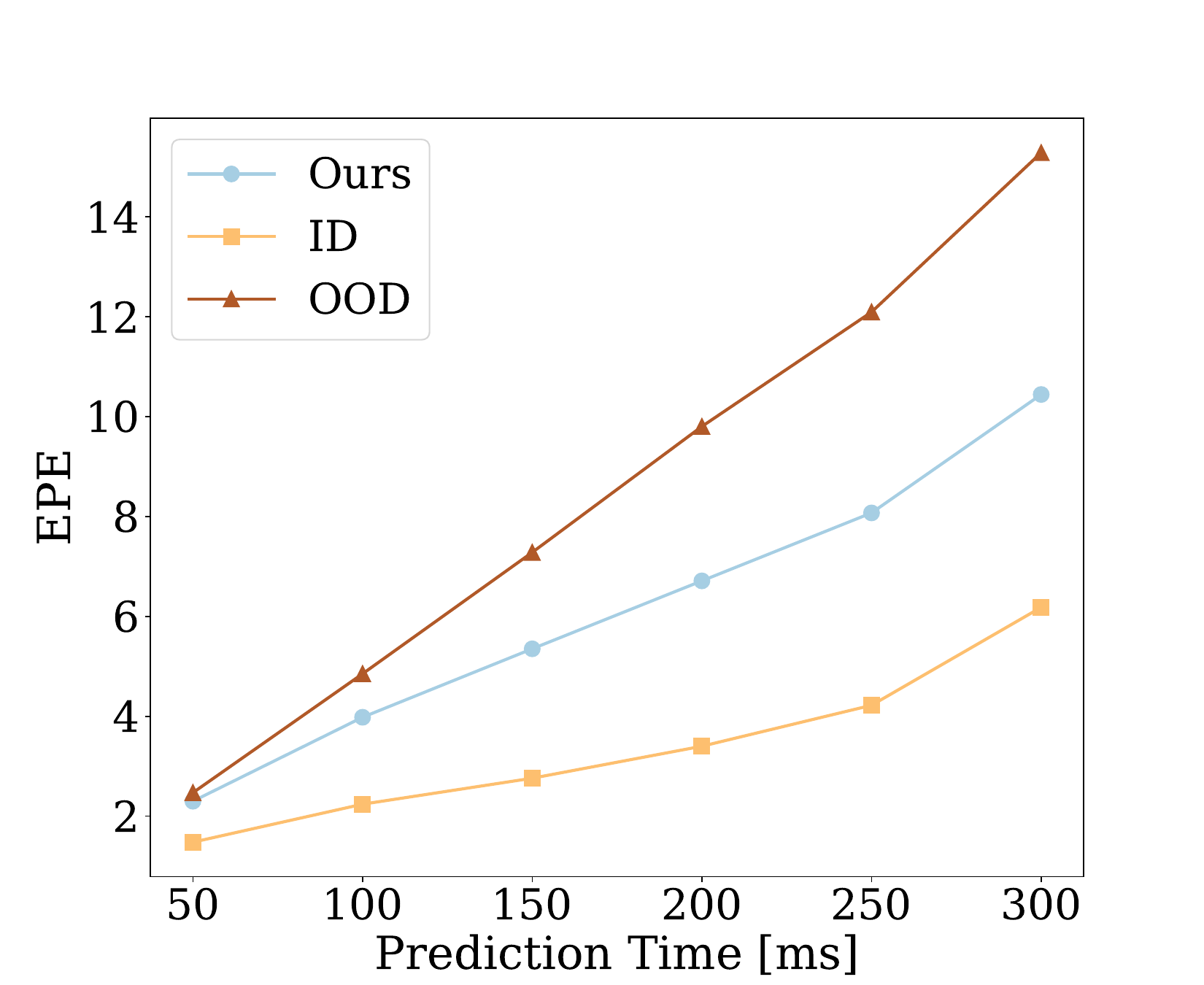}
	\caption{End-point-error vs.~prediction time span for three methods: in-distribution, out-of-distribution and self-supervised (Ours). 
    Using the B\'ezier curve results from \cref{tab:exp:evimo_curve_sensit}.
 }
\label{fig:exp:epe_vs_t}
\end{figure*}
\begin{table}[t]
\centering
\caption{Sensitivity with respect to motion prior on EVIMO2 \cite{Burner22evimo2}. 
\emph{SL, ID}: supervised, in-distribution, 
\emph{SL, OOD}: supervised, out of distribution, 
\emph{SSL}: self-supervised.}
\label{tab:exp:evimo_curve_sensit}
\adjustbox{max width=0.75\linewidth}{
\setlength{\tabcolsep}{6pt}
\begin{tabular}{ll*{3}{S[table-format=2.3]}}
\toprule
 & \textbf{Method} & {\textbf{TEPE} $\downarrow$} & {\textbf{TAE} $\downarrow$} 
 & {\textbf{\%Out} $\downarrow$}
 \\
\midrule

\multirow{3}{*}{\rotatebox[origin=c]{90}{\scalebox{0.8}{SL, ID}}}
 & {BFlow (polyn.)}
        & \unum{1.2}{3.51407885551452} & \unum{2.2}{13.2576541900634} & \unum{1.2}{0.184352248907089} \\
 & {BFlow (learned)}
        & 3.78204464912414 & 13.8688449859619 & 0.189309477806091 \\
 & {BFlow (B\'ezier)}
        & \bnum{3.38006925582885} & \bnum{11.6800241470336} & \bnum{0.166480526328086} \\

\midrule

\multirow{3}{*}{\rotatebox[origin=c]{90}{\scalebox{0.8}{SL, OOD}}}
 & {BFlow (polyn.)}
        & 9.35943126678466 & 20.8827800750732 & 0.362605333328247 \\
 & {BFlow (learned)}
        & 8.66293907165527 & 19.6396293640136 & 0.354855601072311 \\
 & {BFlow (B\'ezier)~\cite{Gehrig24pami}}
        & 8.62916088104248 & 19.9359054565429 & 0.362601906061172 \\

\midrule

\multirow{3}{*}{\rotatebox{90}{\scalebox{0.8}{SSL}}}
& {Ours, polyn.}
        & 6.77561473846435 & 19.7604465484619 & 0.272296130657196 \\
& {Ours, learned}
        & 7.45736408233642 & 19.7804336547851 & 0.284212589263916 \\
& {Ours, B\'ezier}
        & 6.14132022857666 & 16.9782161712646 & 0.254222095012664 \\

\bottomrule
\end{tabular}
}
\end{table}

\subsubsec{EVIMO2 Continuous Flow Dataset (CF-EVIMO2)}. 
MultiFlow~\cite{Gehrig24pami} was first proposed to evaluate long-term ``optical flow'' methods. 
The dataset consists of synthetic videos generated with Internet images as foreground and Flickr30K~\cite{young2014image} images as background. 
Events are generated with a simulator~\cite{Rebecq18corl} by rendering the synthetic scenes at 1000 frames/s. 
However, due to the sim-to-real gap between real and synthetic events, it is challenging to infer the actual performance of the proposed and baseline methods on real datasets. 
Furthermore, Multiflow generates trajectories by pasting 2D foreground objects onto background images, which lacks the challenging cases of self-occlusion due to 3D rotation.

To address these data limitations, we present the \emph{EVIMO2 Continuous Flow Dataset} based on the full 3D data and event data provided by the meticulously designed EVIMO2~\cite{Burner22evimo2} dataset. 
The EVIMO2 GT data provides high-quality 3D scans of the objects, camera poses, object poses, and camera intrinsics, 
which we use to compute optical flow GT as follows.
Given a point $P^i_t$ on the object represented in the camera coordinate system at time $t$, we project it onto a 2D point $p^i_t = \pi(P^i_t)$, where the $\pi(\cdot)$ function represents the perspective projection, including the camera intrinsics. 
The object poses at time $t$ and $t+1$ are provided as $T^{c_{t}}_o$, $T^{c_{t+1}}_o$ in the GT data. 
The rigid-body transformation $T^{c^{t+1}}_{c_t}=(T^{c_{t+1}}_o)^{-1} T^{c_{t}}_o$ maps 3D points in the camera frame at time $t$ to the camera frame at time $t+1$ by aligning the shared object coordinate frame. 
The ``flow'' (or dense motion, since it may not be linear) is defined as $\Delta p^i_{t \rightarrow t+1} \doteq \pi(T^{c^{t+1}}_{c_t} P^i_t) - \pi(P^i_t)$. 
For each sequence, we generate GT dense motion every $10$ms for $300$ms. 
We mask out areas where GT object masks are unavailable. 
Experiments are carried out on the IMO subset of EVIMO2 using the official train and test splits.

\subsubsec{Metrics.}
Ground truth is provided as dense motion from $t = 0$ to several timesteps in increasing order.
We consider $\numSubintervals = 6$ timestamps (i.e., subintervals) in a 300ms window and evaluate the quality of predictions with direct extension of the common optical flow metrics end-point-error (EPE), angular error (AE), and percentage of predicted vectors above a specific EPE threshold (\%Out).
We term the corresponding trajectory metrics TEPE and TAE, where 
\begin{equation}
\textstyle
\text{TEPE} = \frac{1}{\numSubintervals} \sum_{k=1}^{\numSubintervals} \text{EPE}(\Delta\bx_{\text{pred}}(t_k), \Delta\bx_{\text{gt}}(t_k)).
\end{equation}
The number of outliers is calculated on TEPE with a threshold of 3px.

\subsubsec{Implementation Details.}
We use the BFlow network architecture 
and test different motion priors.
Specifically, we report results with polynomial, B\'ezier (with $\numCoeffs = 10$) and learned basis $g_j(t)$ (a small dense neural network with three layers and hidden dimension 64 is trained alongside the main network).

The contrast loss module uses one trajectory per $4 \times 4$ px, and similarly $4 \times 4$~px in the displacement volume.
The weight for the spatial smoothness term is $\lambda = 0.003$, and the number of neighbors in the KNN approach is $\numTraj = 32$.
The loss function $G$ is the $\Lnorm$ norm of the IWE's gradient magnitude \cite{Gallego19cvpr,Shiba24pami}.

\subsubsec{Training Schedule.}
The model is first pre-trained on MultiFlow ($\approx 12000$ samples) for 50 epochs, with a batch size of 10 with an $\Lnorm$ loss on the GT trajectory flow.
We include data augmentation by flipping (horizontal and vertical) and cropping and use the training/test split provided in~\cite{Gehrig24pami}.
These weights are what we refer to as \emph{zero-shot} or \emph{out-of-distribution} (OOD) for the main comparison on EVIMO2.
Afterwards, the pre-trained model is fine-tuned with two different losses on EVIMO2.
Training using our self-supervised loss is carried out for 15 epochs with a batch size of 6.
We refer to this method as \emph{Ours}.
Lastly, a second version is trained directly on the GT flow of EVIMO for 50 epochs, which we refer to as \emph{in-distribution} (ID).
All experiments are performed on Nvidia RTX A6000 GPUs with an AdamW optimizer and a learning rate of $10^{-4}$.

\subsubsec{Baselines.}
This is the first usage of EVIMO2 for dense non-linear flow/motion.
For comparison, we provide additional baselines.
We use ERAFT~\cite{Gehrig21threedv} with the provided DSEC weights to infer linear flow over the whole prediction time.
The timestamps of the intermediate flow are interpolated from the linear prediction.
Additionally, we provide results for the prediction of Paredes et. al.~\cite{Paredes23iccv}.
The network is trained self-supervised on DSEC and performs recurrent prediction steps in time in intervals of 10ms, inferring the total flow at every step by accumulation of the shorter flows.
We unroll this prediction over the whole interval and take the accumulated flow after each 50ms step as the predicted flow.

\subsubsec{Results.}
\Cref{tab:exp:results_evimo} presents the comparison of the three different training modes. The results of our self-supervised loss can improve the zero-shot performance by nearly 30\%.
While direct supervision on EVIMO2 GT data delivers the best results, note that this is only possible because we are comparing on a dataset that was recorded with a motion capture setup and exact object models.
Under real conditions, pixel-level annotations of motion tracks are very difficult to obtain.
Therefore, our loss delivers a promising tool in combination with synthetic pre-training, substantially helping in overcoming the sim-to-real gap.

\Cref{fig:exp:results_evimo2} shows the qualitative comparison of GT tracks and the three training schedules. 
It delivers insights into the failure cases that our method can overcome.
Specifically, the zero-shot model shows errors like overseen object motion (Examples 2 and 3), wrong scale (Ex.~5), and prediction of non-existing motion (Ex.~6), which are improved with self-supervised domain training.

Additionally, \cref{fig:exp:epe_vs_t} shows the error for different prediction time spans.
Intuitively, the error increases for longer intervals.
The visualization shows that our loss improves zero-shot performance, especially for the long prediction times.

\subsubsec{Sensitivity to motion prior.}
\Cref{tab:exp:evimo_curve_sensit} compares our method for different motion priors.
While B\'ezier curves show slightly improved performance over the two basis function methods, self-supervised domain training can improve performance in all cases, which proves the robustness and generality of our method.

\subsection{Results on Optical Flow Estimation}
\label{sec:exp:oflow}

\begin{table*}[t!]
\centering
\caption{Results on the DSEC optical flow benchmark \cite{Gehrig21threedv}.}
\label{tab:exp:dsec}
\adjustbox{max width=\linewidth}{%
\setlength{\tabcolsep}{2pt}

\begin{tabular}{l*{17}{S[table-format=2.3]}}
\toprule
  & 
  & \multicolumn{4}{c}{All}
  & \multicolumn{4}{c}{interlaken\_00\_b}
  & \multicolumn{4}{c}{interlaken\_01\_a}
  & \multicolumn{4}{c}{thun\_01\_a} \\
 \cmidrule(l{1mm}r{1mm}){3-6}
 \cmidrule(l{1mm}r{1mm}){7-10}
 \cmidrule(l{1mm}r{1mm}){11-14}
 \cmidrule(l{1mm}r{1mm}){15-18}

 Method & \text{$t_\text{inf} [\text{ms}]$}
 & \text{EPE $\downarrow$} & \text{AE $\downarrow$} & \text{\%Out $\downarrow$} & \text{FWL $\uparrow$}
 & \text{EPE $\downarrow$} & \text{AE $\downarrow$} & \text{\%Out $\downarrow$} & \text{FWL $\uparrow$}
 & \text{EPE $\downarrow$} & \text{AE $\downarrow$} & \text{\%Out $\downarrow$} & \text{FWL $\uparrow$}
 & \text{EPE $\downarrow$} & \text{AE $\downarrow$} & \text{\%Out $\downarrow$} & \text{FWL $\uparrow$} \\
\midrule

 E-RAFT (SL) \cite{Gehrig21threedv}
 & 46.331
 & \unum{1.2}{0.788} & 10.56 & \unum{1.2}{2.684} & 1.286231422619447
 & \unum{1.2}{1.394} & 6.22 & \unum{1.2}{6.189} & 1.3233361693404235
 & \unum{1.2}{0.899} & 6.881 & \unum{1.2}{3.907} & 1.4233908392179435
 & \unum{1.2}{0.654} & 9.748 & \unum{1.2}{1.87} & 1.200584411 \\

IDNet (SL) \cite{Wu22arxiv}
 &
 & \bnum{0.719} & \bnum{2.723} & \bnum{2.036} & \novalue 
 & \bnum{1.25} & \bnum{2.11} & \bnum{4.353} & \novalue
 & \bnum{0.774} & \bnum{2.25} & \bnum{2.596} & \novalue
 & \bnum{0.572} & \bnum{2.662} & \bnum{1.472} & \novalue \\\midrule

 Paredes et al. (SSL) \cite{Paredes23iccv}
 & \unum{2.2}{40.1}
 & 2.33 & 10.56 & 17.771 & \novalue
 & 3.337 & 6.22 & 25.724 & \novalue
 & 2.489 & 6.881 & 19.153 & \novalue
 & 1.73 & 9.748 & 10.386 & \novalue \\

 RTEF (MB) \cite{Brebion21tits}
 &
 & 4.88 & \novalue & 41.95 & \bnum{2.51}
 & 8.59 & \novalue & 59.84 & \bnum{2.89}
 & 5.94 & \novalue & 47.33 & \bnum{2.92}
 & 3.01 & \novalue & 29.70 & \bnum{2.39}
 \\

EV-FlowNet (SSL)~\cite{Zhu19cvpr}
 &
 & 3.86 & \novalue & 31.45 & 1.30
 & 6.32  & \novalue & 47.95 & 1.46 
 & 4.91 & \novalue & 36.07 & 1.42
 & 2.33 & \novalue & 20.92 & \unum{1.2}{1.32}  \\

MultiCM (MB) \cite{Shiba22eccv}
 & \text{$9.9 \cdot 10^{3}$}
 & 3.472 & 13.983 & 30.855 & 1.365150407655252
 & 5.744 & 9.188 & 38.925 & 1.4992904511327119
 & 3.743 & 9.771 & 31.366 & 1.5137225860194
 & 2.116 & 11.057 & 17.684 & 1.2420968732859203 \\

 Ours (poly, $k=1$)
 & \bnum{7.273}
 & 3.2  & \unum{1.2}{8.53} & 15.21 & \unum{1.2}{1.4603124856948853}
 & 3.21 & \unum{1.2}{4.89}        & 20.45        & \unum{1.2}{1.582527995109558}
 & 2.38 & \unum{1.2}{5.46}        & 17.4        & \unum{1.2}{1.701579213142395}
 & 1.39 & \unum{1.2}{6.99}        & 7.36        & 1.297378420829773 \\

 \\[-0.5ex]
 
\midrule
  & & \multicolumn{4}{c}{thun\_01\_b}
  & \multicolumn{4}{c}{zurich\_city\_12\_a}
  & \multicolumn{4}{c}{zurich\_city\_14\_c}
  & \multicolumn{4}{c}{zurich\_city\_15\_a} \\

 \cmidrule(l{1mm}r{1mm}){3-6}
 \cmidrule(l{1mm}r{1mm}){7-10}
 \cmidrule(l{1mm}r{1mm}){11-14}
 \cmidrule(l{1mm}r{1mm}){15-18}
 
 Method & & \text{EPE $\downarrow$} & \text{AE $\downarrow$} & \text{\%Out $\downarrow$} & \text{FWL $\uparrow$}
 & \text{EPE $\downarrow$} & \text{AE $\downarrow$} & \text{\%Out $\downarrow$} & \text{FWL $\uparrow$}
 & \text{EPE $\downarrow$} & \text{AE $\downarrow$} & \text{\%Out $\downarrow$} & \text{FWL $\uparrow$}
 & \text{EPE $\downarrow$} & \text{AE $\downarrow$} & \text{\%Out $\downarrow$} & \text{FWL $\uparrow$} \\
\midrule

 E-RAFT (SL) \cite{Gehrig21threedv}
 & & \unum{1.2}{0.578} & 8.409 & \unum{1.2}{1.518} & 1.1767931182449056
 & \unum{1.2}{0.612} & 23.164 & \bnum{1.057} & 1.1161122798926002
 & \bnum{0.713} & 10.226 & \bnum{1.913} & 1.4688118
 & \unum{1.2}{0.589} & 8.878 & \unum{1.2}{1.303} & 1.335840906676725 \\

 IDNet (SL) \cite{Wu22arxiv}
 & & \bnum{0.546} & \bnum{2.07} & \bnum{1.35} & \novalue 
 & \bnum{0.603} & \bnum{4.556} & \unum{1.2}{1.161} & \novalue
 & 0.76 & \bnum{3.742} & 2.735 & \novalue
 & \bnum{0.548} & \bnum{2.545} & \bnum{1.022} & \novalue \\\midrule

 Paredes et al. (SSL) \cite{Paredes23iccv}
 & & 1.657 & 8.409 & 9.343 & \novalue
 & 2.724 & 23.164 & 26.649 & \novalue
 & 2.635 & 10.226 & 23.005 & \novalue
 & 1.686 & 8.878 & 9.977 & \novalue \\

 RTEF (MB) \cite{Brebion21tits}
 & & 3.91 & \novalue & 34.69 & \bnum{2.48}
 & 3.14 & \novalue & 34.08 & \bnum{1.42}
 & 4.00 & \novalue & 45.67 & \bnum{2.67}
 & 3.78 & \novalue & 37.99 & \bnum{2.82}
 \\

 EV-FlowNet (SSL)~\cite{Zhu19cvpr}
 & & 3.04 & \novalue & 25.41 & 1.33
 & 2.62 & \novalue & 25.80 & 1.03
 & 3.36 & \novalue & 36.34 & 1.24
 & 2.97 & \novalue & 25.53 & 1.33 \\

 MultiCM (MB) \cite{Shiba22eccv}
 & & 2.48 & 12.045 & 23.564 & 1.24194368901993
 & 3.862 & 28.613 & 43.961 & \unum{1.2}{1.1375111350019202}
 & 2.724 & 12.624 & 30.53 & 1.4985924122973489
 & 2.347 & 11.815 & 20.987 & 1.4119389239425226 \\

Ours (poly, $k=1$)
 & & 1.54 & \unum{1.2}{6.55}  & 9.69   & \unum{1.2}{1.3275487422943115}
 & 8.33 & \unum{2.2}{20.16} & 22.39  & 1.1288989782333374
 & 1.78 & \unum{1.2}{8.79}  & 12.99  &  \unum{1.2}{1.5562018156051636}
 & 1.45 & \unum{1.2}{6.27}  & 8.34   & \unum{1.2}{1.5112605094909668} \\

\bottomrule
\end{tabular}
}
\end{table*}

\subsubsec{Dataset, Metrics and  Details.}
We compare our method on the DSEC optical flow benchmark \cite{Gehrig21threedv}. 
The dataset consists of sequences from a Prophesee Gen3 event camera, with a resolution of $640 \times 480$ px, mounted on a driving car.
Ground truth is calculated as the motion field from a co-deployed depth sensor.

We compare the common flow metrics EPE, AE, and percentage of outliers (\%Out), where an outlier is a flow vector with $\text{EPE} > 3$px.
Additionally, we evaluate the Flow Warp Loss (FWL)~\cite{Stoffregen20eccv}.
$\text{FWL}>1$ means that the IWE is sharper than pixel-wise event accumulation (i.e., IWE with zero optical flow).

We use a simple U-Net~\cite{Ronneberger15icmicci} as the backbone in the linear flow experiments (similar architecture like EV-FlowNet~\cite{Zhu18rss}) and choose polynomial basis $g_j(t) = t^j$ of degree 1, which effectively leads to a linear motion prior.
We use $\numTraj = 32$ nearest neighbors and $\numBins = 15$ time bins in the displacement map.
The weight for the regularizer is $\lambda = 0.003$.
The network trains for 50 epochs with and Adam optimizer, a learning rate of $10^{-4}$, and a total batch size of 28 on two RTX A6000.

\subsubsec{Results.}
\Cref{tab:exp:dsec} shows the DSEC benchmark results, confirming that our model can generate high-quality optical flow results without having access to any GT.
It furthermore achieves the best results among all contrast-maximization--based methods on six out of seven sequences, and only methods directly supervised on GT perform better.
Within the self-supervised methods, it improves the AE by 19\% and the number of inliers by 14\%.
The only exception is the night sequence \emph{zurich\_city\_14}.
Contrast maximization is based on the brightness constancy assumption, which is violated here by flickering street lights.
Hence, in these areas, the contrast loss does not provide reasonable flow predictions.
While our methods still provide a better AE and \%Out at night, higher outliers in the flickering regions prevent the difference in the average EPE from reflecting the improved performance of our method on the other six sequences.

\Cref{fig:exp:results_dsec} shows qualitative results of our method.
In comparison to another self-supervised optical flow method (d), the visualizations are noticeably sharper, allowing for a more precise representation of the motion in the scene.
Moreover, the model performs well in delineating the contours of foreground objects, without the over-smoothing effect often observable in flow methods.
The IWE (\cref{fig:exp:results_dsec}b) shows sharp results, aligning with the overall high FWL values reported in \cref{tab:exp:dsec}.
While the FWL metric also increases in event collapse (e.g., for RTEF \cite{Brebion21tits}), the IWE visualization reveals that here the model is well-regularized, confirming that the predicted motion aligns the events correctly.
The third example in \cref{fig:exp:results_dsec} highlights how our model has improved predictions with a single forward pass, where models relying on temporal recurrence (see (d)) fail at the beginning and need a warm-up stage resulting in higher latency.

\subsubsec{Inference time.}
Additionally, \cref{tab:exp:dsec} provides inference times $t_\text{inf}$ for several methods.
Note that our method is about $5\times$ faster than the competitive baselines.
The reason is that we do not rely on any recurrence in this optical flow application, such as recurrence in time or RAFT-inspired refinement steps.

\def\figWidth{0.186\linewidth}
\begin{figure*}[t]
	\centering
    {\scriptsize
    \setlength{\tabcolsep}{1pt}
	\begin{tabular}{
	>{\centering\arraybackslash}m{0.3cm} 
	>{\centering\arraybackslash}m{\figWidth} 
	>{\centering\arraybackslash}m{\figWidth} 
	>{\centering\arraybackslash}m{\figWidth} 
	>{\centering\arraybackslash}m{\figWidth} 
	>{\centering\arraybackslash}m{\figWidth}
    }
        \rotatebox{90}{\makecell{\tiny zurich city 15a}}
		&\gframe{\includegraphics[clip,trim={0cm 0cm 0cm 0cm},width=\linewidth]{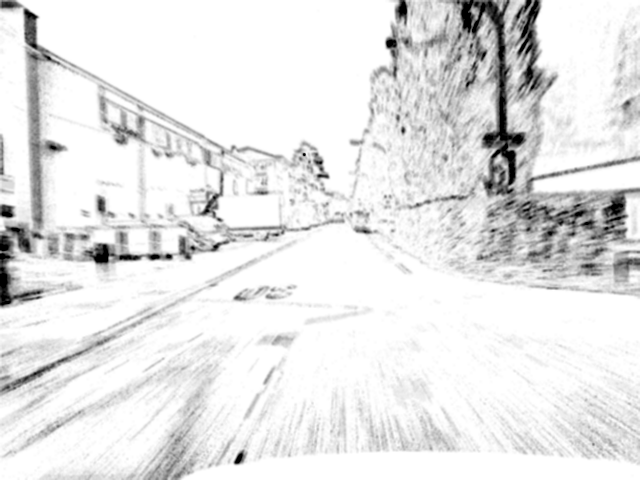}}
		&\gframe{\includegraphics[clip,trim={0cm 0cm 0cm 0cm},width=\linewidth]{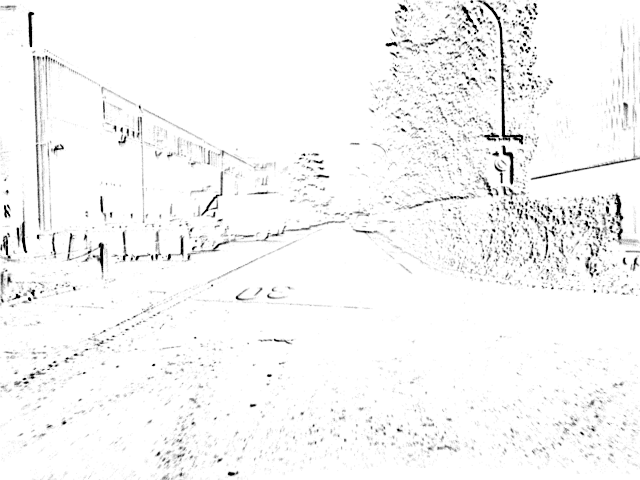}}
		&\includegraphics[clip,trim={0cm 0cm 0cm 0cm},width=\linewidth]{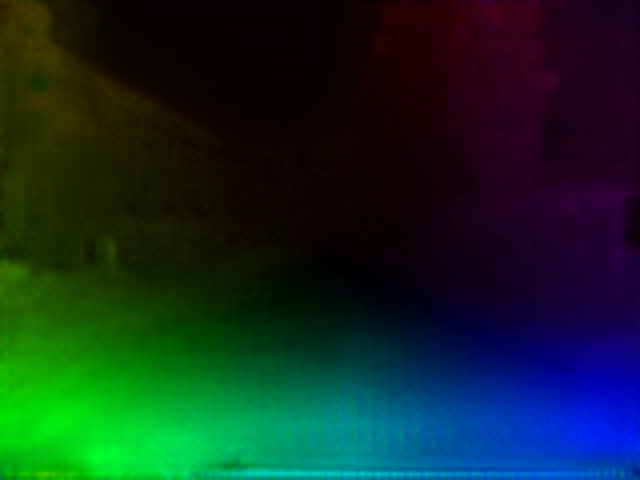}
		&\includegraphics[clip,trim={0cm 0cm 0cm 0cm},width=\linewidth]{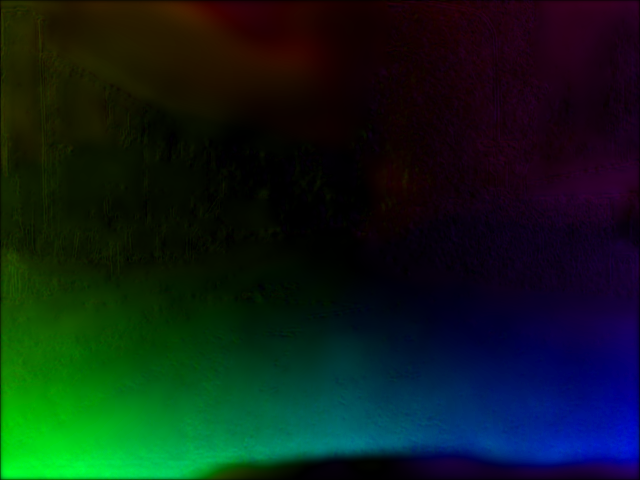}
        &\includegraphics[clip,trim={0cm 0cm 0cm 0cm},width=\linewidth]{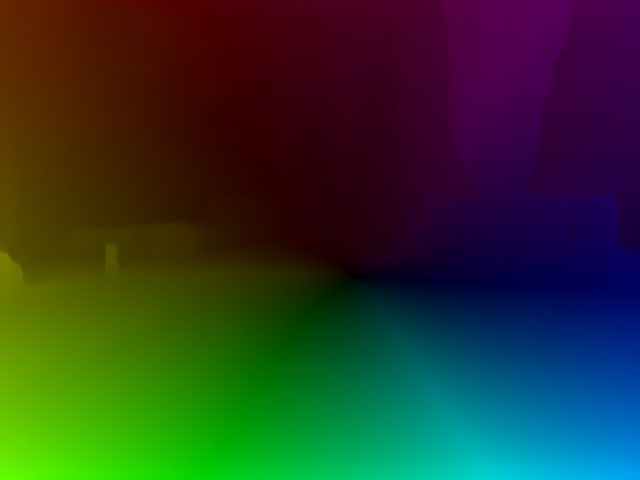} \\

        \rotatebox{90}{\makecell{\tiny zurich city 14c}}
		&\gframe{\includegraphics[clip,trim={0cm 0cm 0cm 0cm},width=\linewidth]{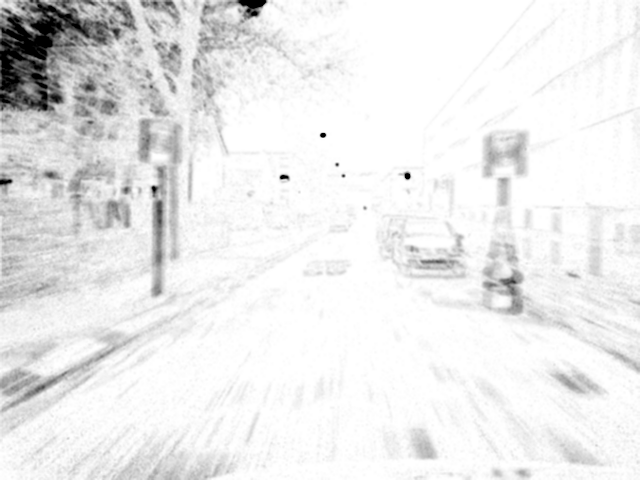}}
		&\gframe{\includegraphics[clip,trim={0cm 0cm 0cm 0cm},width=\linewidth]{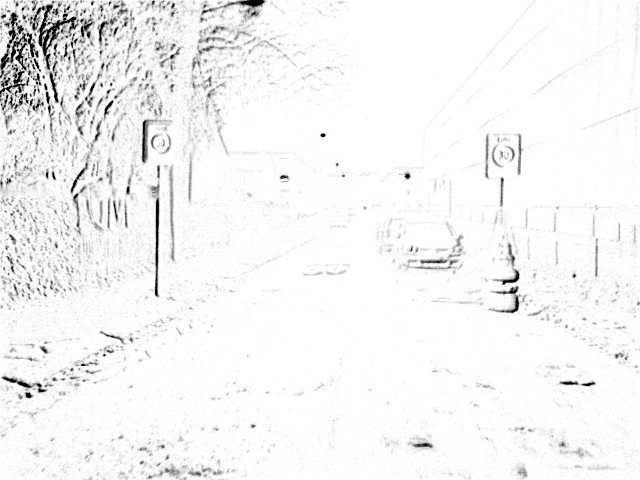}}
		&\includegraphics[clip,trim={0cm 0cm 0cm 0cm},width=\linewidth]{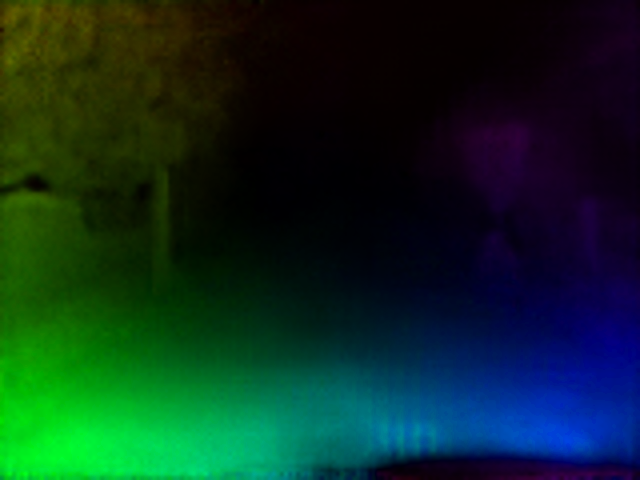}
		&\includegraphics[clip,trim={0cm 0cm 0cm 0cm},width=\linewidth]{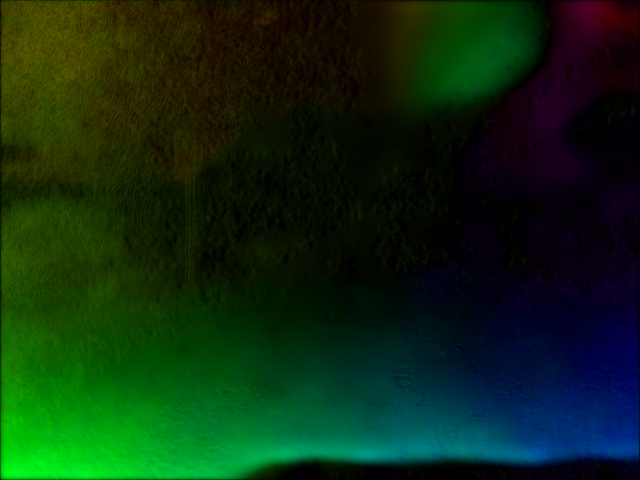}
        &\includegraphics[clip,trim={0cm 0cm 0cm 0cm},width=\linewidth]{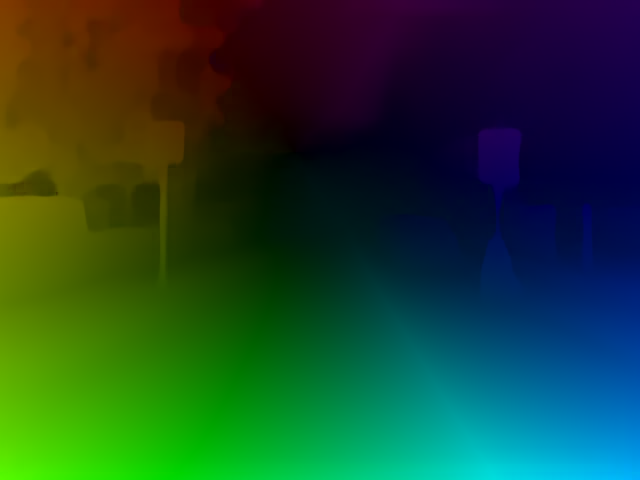} \\

        \rotatebox{90}{\makecell{\tiny interlaken 01a}}
		&\gframe{\includegraphics[clip,trim={0cm 0cm 0cm 0cm},width=\linewidth]{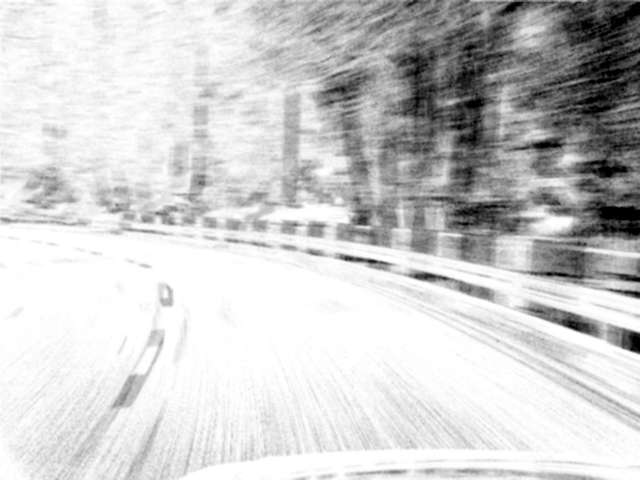}}
		&\gframe{\includegraphics[clip,trim={0cm 0cm 0cm 0cm},width=\linewidth]{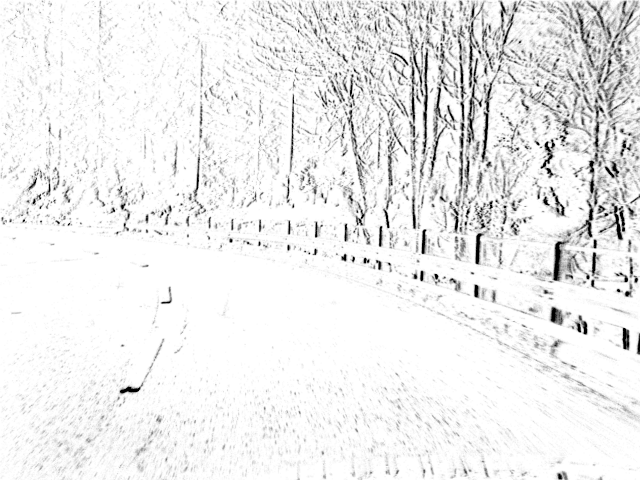}}
		&\includegraphics[clip,trim={0cm 0cm 0cm 0cm},width=\linewidth]{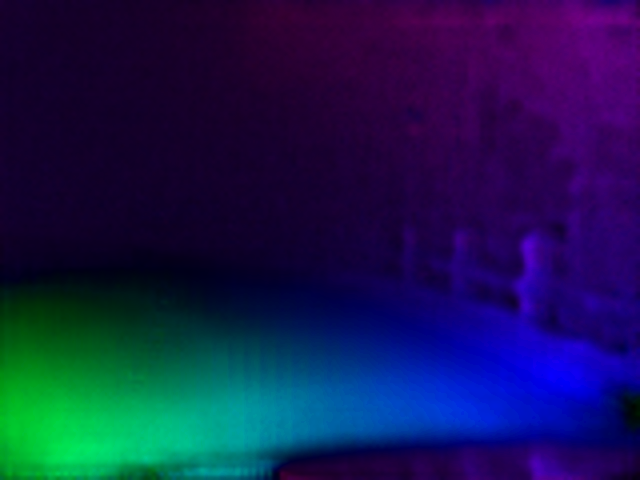}
		&\includegraphics[clip,trim={0cm 0cm 0cm 0cm},width=\linewidth]{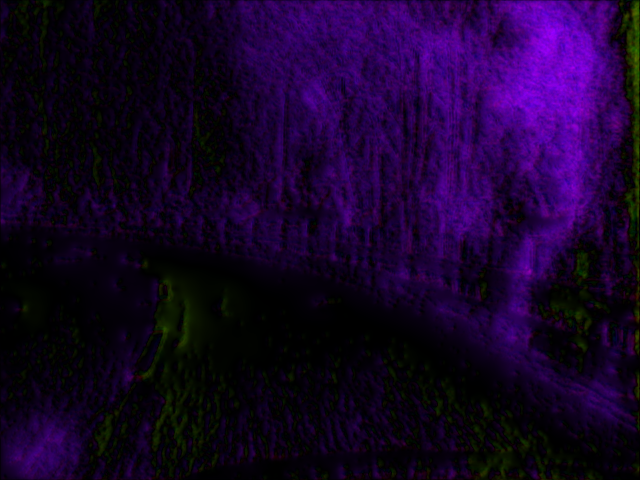}
        &\includegraphics[clip,trim={0cm 0cm 0cm 0cm},width=\linewidth]{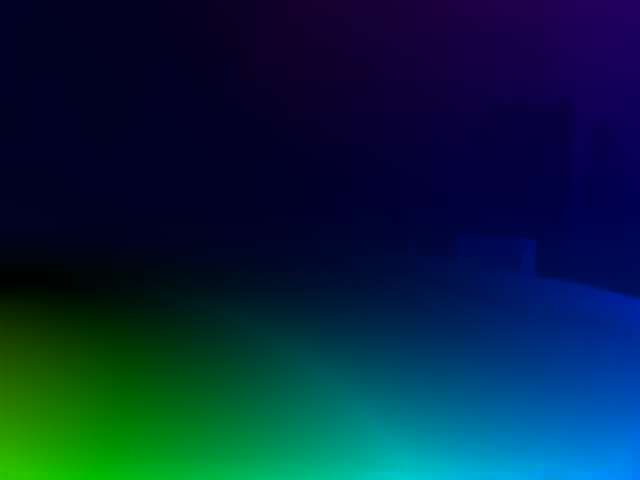} \\

		& \textbf{(a)} Events
		& \textbf{(b)} IWE (Ours)
		& \textbf{(c)} Flow (Ours)
		& \textbf{(d)} Flow Paredes~\cite{Paredes23iccv}
        & \textbf{(e)} Flow ERAFT (SL)~\cite{Gehrig21threedv}
	\end{tabular}
	}
	\caption{Results on DSEC. Image of warped events and predicted flow by three methods.}
	\label{fig:exp:results_dsec}
\end{figure*}

\subsubsec{Ablation and Sensitivity.}
\Cref{tab:exp:ablation_dsec} lists the performance under different loss settings.
It confirms that \emph{the number of neighboring trajectories $\numTraj$} in the KNN approach has a regularizing effect and is important for good performance.
The decreased performance on the benchmark for a lower $\numTraj$ aligns with the visual impression in \cref{fig:exp:sensitivity_knn}: 
models with lower $\numTraj$ show artifacts and a stronger susceptibility to the aperture problem, as is observable on the rock pattern.

\Cref{tab:exp:ablation_dsec} confirms that the \emph{multi-reference approach} is crucial for regularizing the contrast loss.
While using three instead of one $\tref$ improves performance, our approach using a randomized $\tref$ leads to an additional performance boost.
Lastly, the non-linear prior showed no performance improvement on DSEC.

\begin{table}[t]
\centering
\caption{
Sensitivity and Ablation Study for DSEC.
Ours corresponds to ``Ours'' in \cref{tab:exp:dsec}.
Configurations marked with ``{--}'' are unchanged from our main result.
$\Delta$ specifies the change with respect to the original configuration.
}
\label{tab:exp:ablation_dsec}
\adjustbox{max width=\linewidth}{
\setlength{\tabcolsep}{2pt}
\begin{tabular}{ll*{3}{S[table-format=2]}l*{6}{S[table-format=2.3]}}
\toprule
&
 & $\boldsymbol{\numTraj}$  &  $\boldsymbol{N_\text{tref}}$ & $\boldsymbol{\numCoeffs}$  & \textbf{\text{Motion prior}}
 & {\textbf{EPE} $\downarrow$} & {$\boldsymbol{\Delta}$\textbf{EPE}} & {\textbf{AE} $\downarrow$} & {$\boldsymbol{\Delta}$\textbf{AE}}
 & {\textbf{\%Out} $\downarrow$} & {$\boldsymbol{\Delta}$\textbf{\%Out}}
 \\
\midrule

& {Ours}
    & \text{$32$}   & \text{$\sim \mathcal{U}(0, 1)$} & \text{$1$} & polynomial
    & \bnum{3.2}   & 
    & \bnum{8.53}  & 
    & \bnum{15.21} & \\

\midrule

& Number of neighbor
    & \text{1} & {--} & {--} & {--}
    & 4.576 & 1.07
    & 13.732 & 5.2
    & 27.902 & 12.69 \\

& trajectories
    & \text{8} & {--} & {--} & {--}
    & 3.512  & 0.31
    & 13.199 & 4.67
    & 23.388 & 8.178 \\

\midrule

& Number of 
    & {--} & \text{3} & {--} & {--}
    & 4.46 & 1.26
    & 14.434 & 5.904
    & 28.602 & 13.39 \\

& reference times
    & {--} & \text{1} & {--} & {--}
    & 7.264 & 4.06
    & 18.719 & 10.19
    & 44.751 & 29.54 \\

\midrule

& Type and degree
    & {--} & {--} & \text{5} & learned
    & 3.22 & 0.02
    & 8.59 & 0.06
    & 15.34 & 0.13 \\

& of motion prior
    & {--} & {--} & \text{5} & polynomial
    & 3.28 & 0.06
    & 8.61 & 0.08 &
    15.68 & 0.47 \\

\bottomrule
\end{tabular}
}
\end{table}

\def\figWidth{0.235\linewidth}
\begin{figure*}[ht!]
	\centering
    {\scriptsize
    \setlength{\tabcolsep}{1pt}
	\begin{tabular}{
	>{\centering\arraybackslash}m{\figWidth} 
	>{\centering\arraybackslash}m{\figWidth} 
    >{\centering\arraybackslash}m{0.2cm}
	  >{\centering\arraybackslash}m{\figWidth} 
	  >{\centering\arraybackslash}m{\figWidth} 
    }
		\gframe{\includegraphics[clip,trim={0cm 0cm 0cm 0cm},width=\linewidth]{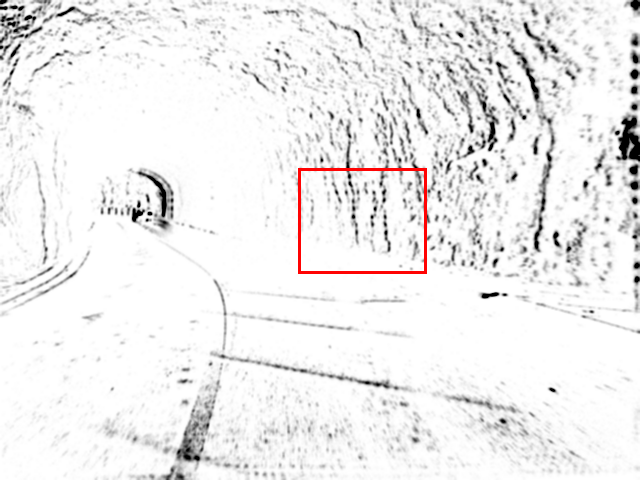}}
		&\gframe{\includegraphics[clip,trim={7cm 5cm 5cm 4cm},width=\linewidth]{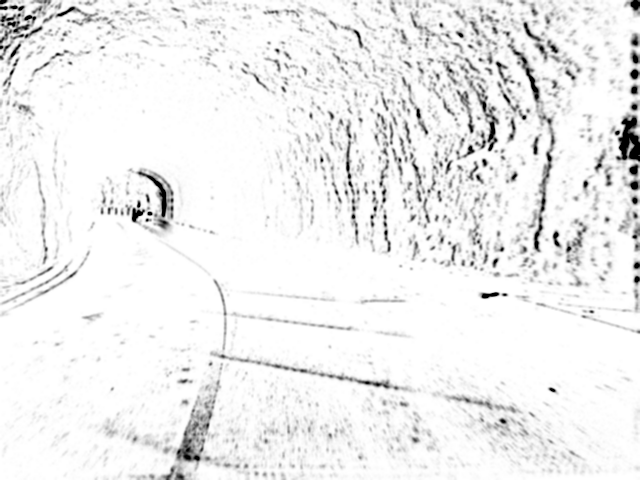}}
        & \;
		&\gframe{\includegraphics[clip,trim={0cm 0cm 0cm 0cm},width=\linewidth]{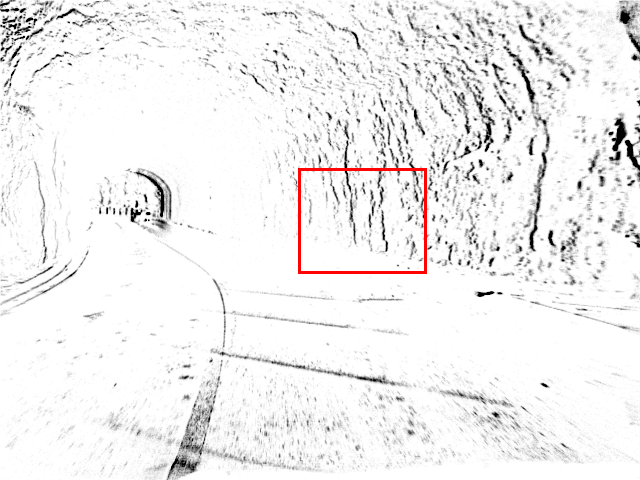}}
		&\gframe{\includegraphics[clip,trim={7cm 5cm 5cm 4cm},width=\linewidth]{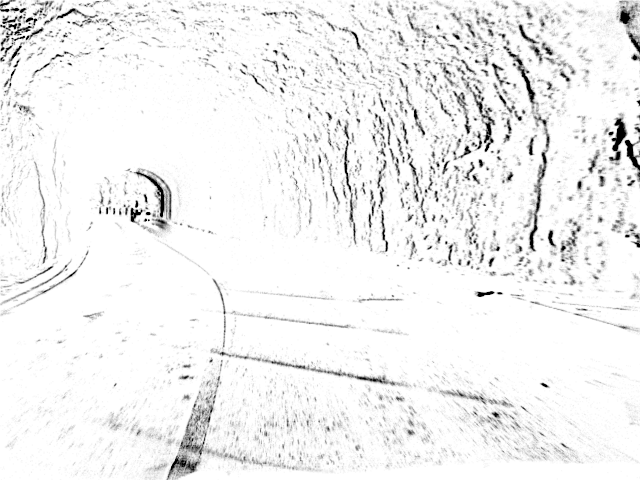}} \\

        \multicolumn{2}{c}{$\numTraj = 1$ (some event collapse)}
        & 
		& \multicolumn{2}{c}{$\numTraj = 32$ (Ours)}
	\end{tabular}
	}
	\caption{Visual effect of the number of neighbors $\numTraj$ on the predicted flow.}
	\label{fig:exp:sensitivity_knn}
\end{figure*}

\section{Limitations}

Like all CM-based methods, ours is based on the brightness constancy assumption.
Therefore, it shows limitations in estimating flow from events that are not caused by motion, e.g., from flickering lights.

Our KNN interpolation shows a good trade-off between granularity and regularization; nevertheless, our method is limited by the aperture problem inherent to optical flow.

Lastly, during training, all raw events are passed to the loss module, increasing training time.
Furthermore, gradient calculation is performed through the additional step of event warping.
These steps increase memory requirements and training time of our loss compared to supervised methods.

\section{Conclusion}
\label{sec:conclusion}

We introduced a new loss formulation based on the contrast maximization framework, by combining it with a non-linear trajectory prior.
It is a versatile tool that works with a variety of model architectures and trajectory representations.
We presented an efficient method to make the high-dimensional assignment between millions of events and thousands of trajectories feasible.
Our experiments show clear advantages against supervised methods on unseen data, where under real-world circumstances no GT is available.
Additionally, a U-Net trained with our loss shows state-of-the-art performance on the DSEC optical flow benchmark, while being substantially faster than the previously best methods.

\section*{Acknowledgments}
We thank Dr.~Cornelia Ferm\"uller and the NeuroPAC network for fostering collaborations within the event-based community (NSF OISE 2020624).
Funded by the Deutsche Forschungsgemeinschaft (DFG, German Research Foundation) under Germany’s Excellence Strategy -- EXC 2002/1 ``Science of Intelligence'' -- project number 390523135.
We furthermore gratefully acknowledge the support by the following grants: NSF FRR 2220868, NSF IIS-RI 2212433, NSF TRIPODS 1934960, ONR N00014-22-1-2677.

\section*{Supplementary}

\subsection*{Loss Details}
Our proposed loss function furthermore has the following details, which we experimentally found to have a slight advantage over not using them.

\subsubsec{Scaling warped events by time to $\tref$.}
The contribution of each warped event to the IWE is weighted by its distance $\Delta t = |\tref - t_k|$.

\subsubsec{Masking image border.}
We mask events transported outside the image plane by the warp.
They do not contribute to the loss.

\subsubsec{Polarity-split IWEs.}
We treat positive and negative events separately when calculating the IWE.
Effectively, at each iteration, two IWEs are evaluated.

\subsection*{Results on MultiFlow dataset}
\Cref{tab:supp:results_mf} shows results on the MultiFlow dataset \cite{Gehrig24pami}, which are coherent with the results on EVIMO.
Our self-supervised method performs second best after models directly supervised on the ground truth trajectories.
Similarly, B\'ezier curves have a slight performance advantage, over the other tested trajectories.
A model trained with additional frames as input shows an improved performance over the event-only model showing potential for multi-domain extensions.

\begin{table}[ht!]
\centering
\caption{
Results on MultiFlow dataset~\cite{Gehrig24pami}.
``Frames'' indicates whether frames at $t_s$ and $t_e$ were used as additional input to the artificial neural network or not.
}
\label{tab:supp:results_mf}
\adjustbox{max width=\linewidth}{
\setlength{\tabcolsep}{3pt}
\begin{tabular}{ll*{4}{S[table-format=2.4]}}
\toprule
& \textbf{Method} & \textbf{\text{Frames}} & {\textbf{TEPE} $\downarrow$} & {\textbf{TAE} $\downarrow$} 
& {\textbf{\%Out} $\downarrow$}
\\
\midrule

\multirow{3}{*}{\rotatebox[origin=c]{90}{SL}}
& {BFlow, polyn.}
       & \xmark & 1.70905399322509 & 5.92629098892211 & 0.0925752148032188 \\
& {BFlow, B\'ezier~\cite{Gehrig24pami}}
       & \xmark & 1.67861235141754 & 5.86530351638793 & 0.0878454148769378 \\
\midrule

\multirow{4}{*}{\rotatebox{90}{SSL}}
& {Paredes et. al~\cite{Paredes23iccv}}
        & \xmark & 14.8137 & 61.1371 & 0.8392 \\
& {Ours, polyn.}
       & \xmark & 8.5691827374651 & 31.37957133854 & 0.476549012235 \\
& {Ours, B\'ezier}
       & \xmark & 8.14813804626464 & 29.8925838470459 & 0.459626883268356 \\

& {Ours, B\'ezier}
       & \cmark & 7.2698 & 27.7564392015743 & 0.430123457487288 \\

\bottomrule
\end{tabular}
}
\end{table}

\subsection*{Runtime of Submodules}
The first column of Tab.~4 in the main paper reports inference times of different methods.
Here is a breakdown of module times for our method.
network inference (Main Fig. 2c)): 7.27ms,
computation of trajectories (Fig. 2c to 2d): 0.19ms.
For the loss module (Main Fig. 2e), used only at training time:
flow interpolation (Main Fig. 2e1 to 2e3): 86ms,
event warping (Main Fig. 2e4): 7.18ms,
building IWE (Main Fig. 2e5): 8.34ms
Notably, the flow interpolation is the slowest step but is only required during training, not during inference.

\subsection*{Ablations on EVIMO2}
\Cref{tab:supp:evimo_ablations} shows additional ablations on EVIMO2 (analogue to the tests on DSEC in Tab.~5 of the paper).
The results confirm most design choices, like the number of neighbors, and the use of a randomized reference time.
As the prediction time is longer in this dataset, we also see the influence of the motion prior type and degree.
We found B\'ezier curves with degree $\numCoeffs = 10$ to work best.
\begin{table}[ht]
\centering
\caption{Sensitivity and ablation study on EVIMO2 data.
Ours corresponds to ``Ours, B\'ezier'' in Tab.~3 of the main paper.
Configurations marked with ``{--}'' are unchanged from our main result.}
\label{tab:supp:evimo_ablations}
\adjustbox{max width=\linewidth}{
\setlength{\tabcolsep}{3pt}
\begin{tabular}{lccccccc}
\toprule
 & $\boldsymbol{\numTraj}$  &  $\boldsymbol{N_\text{tref}}$ & $\boldsymbol{\numCoeffs}$  & \textbf{\text{Motion prior}}
 & {\textbf{TEPE} $\downarrow$} & {\textbf{TAE}$\downarrow$} & {\textbf{\%Out} $\downarrow$} \\
  \midrule
 Ours & 32 & \text{$\sim \mathcal{U}(0, 1)$} & 10 &  B\'ezier & 6.14 & 16.98  & 0.25 \\
      & 8 & {--} & {--} &  {--} & 6.63 & 18.11  & 0.26 \\
      & 64 & {--} & {--} &  {--} & 6.61 & 19.18  & 0.29 \\
      & {--} & 1 & {--} &  {--} & 6.76 & 18.32  & 0.26 \\
      & {--} & {--} & 5 &  {--} & 6.25 & 17.81  & 0.25 \\
      & {--} & {--} & 30 &  {--} & 6.44 & 17.95  & 0.28 \\
      & {--} & {--} & 1 &  polynomial & 7.97 & 21.98  & 0.39 \\
 \bottomrule
\end{tabular}
}
\end{table}

\subsection*{Results on MVSEC}
\begin{table*}[!t]
\centering
\caption{Quantitative evaluation on MVSEC data \cite{Zhu18ral}. 
Best in bold, runner-up underlined. 
SL: supervised learning; SSL$_{\text{F}}$: SSL trained with grayscale images; SSL$_{\text{E}}$: SSL trained with events; MB: model-based methods.}
\label{tab:mvsec}
\adjustbox{max width=\linewidth}{
\setlength{\tabcolsep}{3pt}
	\begin{tabular}{lcccccccc}
		\toprule
		  \multirow{2}{*}{} & \multicolumn{2}{c}{indoor\_flying1} && \multicolumn{2}{c}{indoor\_flying2} && \multicolumn{2}{c}{indoor\_flying3} \\\cline{2-3}\cline{5-6}\cline{8-9}
		  & EPE$\downarrow$& $\%_{\text{3PE}}$$\downarrow$&& EPE$\downarrow$& $\%_{\text{3PE}}$$\downarrow$&& EPE$\downarrow$& $\%_{\text{3PE}}$$\downarrow$\\
        \midrule
		\parbox[t]{0mm}{\multirow{5}{*}{\rotatebox[origin=c]{90}{SL}}}
		\hspace{10pt}EV-FlowNet+ \cite{Stoffregen20eccv} & 0.56 & 1.00 && \underline{0.66} & \underline{1.00} && \underline{0.59} & 1.00\\
		\hspace{12.5pt}E-RAFT \cite{Gehrig21threedv} & - & - && - & - && - & - \\
		\hspace{12.5pt}EV-FlowNet \cite{Gehrig21threedv} & - & - && - & - && - & - \\
		\hspace{12.5pt}TMA \cite{Liu23iccv} & 1.06 & 3.63 && 1.81 & 27.29 && 1.58 & 23.26 \\
		\hspace{12.5pt}Cuadrado \textit{et al}.\ \cite{Cuadrado23fns} & 0.58 & - && 0.72 & - && 0.67 & - \\
		\midrule
		\parbox[t]{0mm}{\multirow{2}{*}{\rotatebox[origin=c]{90}{SSL$_{\text{F}}$}}}
		\hspace{10pt}EV-FlowNet \cite{Zhu18rss} & 1.03 & 2.20 && 1.72 & 15.1 && 1.53 & 11.9\\
		\hspace{12.5pt}Ziluo \textit{et al}.\ \cite{Ding22aaai} & 0.57 & 0.10 && 0.79 & 1.60 && 0.72 & 1.30 \\
		\midrule
		\parbox[t]{0mm}{\multirow{5}{*}{\rotatebox[origin=c]{90}{SSL$_{\text{E}}$}}}
		\hspace{10pt}EV-FlowNet \cite{Zhu19cvpr} & 0.58 & \textbf{0.00} && 1.02 & 4.00 && 0.87 & 3.00 \\
		\hspace{12.5pt}EV-FlowNet \cite{Paredes21cvpr} & 0.79 & 1.20 && 1.40 & 10.9 && 1.18 & 7.40 \\
		\hspace{12.5pt}EV-FlowNet \cite{Shiba22eccv} & - & - && - & - && - & -\\
		\hspace{12.5pt}ConvGRU-EV-FlowNet \cite{Hagenaars21neurips} & 0.60 & 0.51 && 1.17 & 8.06 && 0.93 & 5.64 \\
		\hspace{12.5pt}Paredes \textit{et al.} \cite{Paredes23iccv} & \underline{0.44} & \textbf{0.00} && 0.88 & 4.51 && 0.70 & 2.41 \\
        \hspace{12.5pt}\textbf{Ours} & 0.45  & \underline{0.09} && 0.71 & 2.40 && 0.6 & \underline{0.93} \\
		\midrule
		\parbox[t]{0mm}{\multirow{3}{*}{\rotatebox[origin=c]{90}{MB}}}
		\hspace{10pt}Akolkar \textit{et al}.\ \cite{Akolkar22pami} & 1.52 & - && 1.59 & - && 1.89 & - \\
		\hspace{12.5pt}Brebion \textit{et al}.\ \cite{Brebion21tits} & 0.52 & 0.10 && 0.98 & 5.50 && 0.71 & 2.10\\
		\hspace{12.5pt}Shiba \textit{et al}.\ \cite{Shiba22eccv} & \textbf{0.42} & \underline{0.09} && \textbf{0.60} & \textbf{0.59} && \textbf{0.50} & \textbf{0.29}\\
		\bottomrule
    \end{tabular}
    }%
\end{table*}

For completeness, \cref{tab:mvsec} provides a qualitative comparison of our method with state-of-the-art techniques on MVSEC data \cite{Zhu18ral}. 
The input for a sample consists of all event data between two consecutive frames 
(GT depth from the LiDAR is temporally upsampled to the frame rate of the DAVIS346 event cameras used, at 45Hz \cite{Zhu18rss,Shiba24pami}).
The models were trained with the same hyperparameters as reported for the DSEC dataset.
The results confirm the good performance of our model and outperforms most baseline methods.
It performs on average 14\% better than Paredes et al.~\cite{Paredes23iccv} and 13\% worse than the test-time optimization-based method from Shiba et al.~\cite{Shiba22eccv}.

\clearpage
\subsection{Additional Qualitative Results on EVIMO2 and DSEC}
\def\figWidth{0.2\linewidth}
\begin{figure*}[ht!]
	\centering
    {\scriptsize
    \setlength{\tabcolsep}{2pt}
	\begin{tabular}{
	>{\centering\arraybackslash}m{0.3cm} 
	>{\centering\arraybackslash}m{\figWidth} 
	>{\centering\arraybackslash}m{\figWidth} 
	>{\centering\arraybackslash}m{\figWidth} 
	>{\centering\arraybackslash}m{\figWidth} 
    }

        \rotatebox{90}{\makecell{Example 7}}
		&\gframe{\includegraphics[clip,trim={0cm 0cm 0cm 0cm},width=\linewidth]{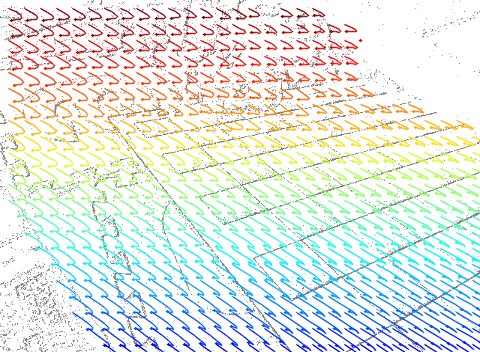}}
        &\gframe{\includegraphics[clip,trim={0cm 0cm 0cm 0cm},width=\linewidth]{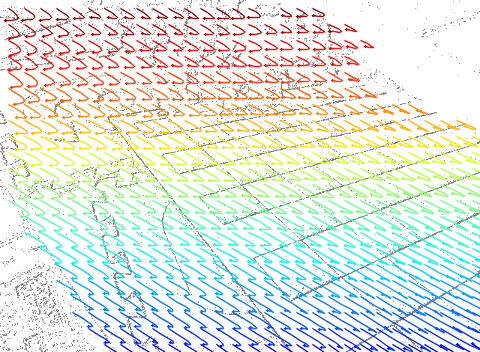}}
		&\gframe{\includegraphics[clip,trim={0cm 0cm 0cm 0cm},width=\linewidth]{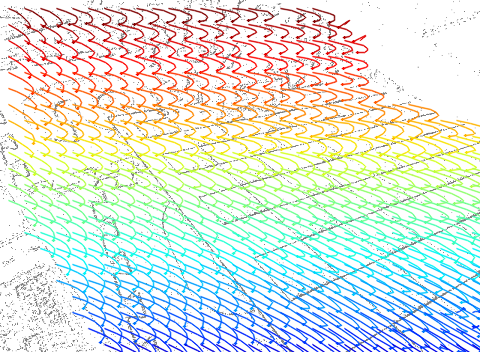}}
        &\gframe{\includegraphics[clip,trim={0cm 0cm 0cm 0cm},width=\linewidth]{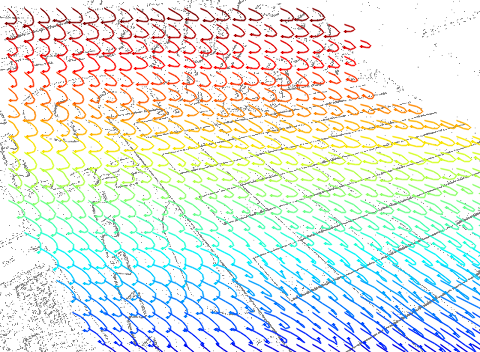}} \\

        \rotatebox{90}{\makecell{Example 8}}
		&\gframe{\includegraphics[clip,trim={0cm 0cm 0cm 0cm},width=\linewidth]{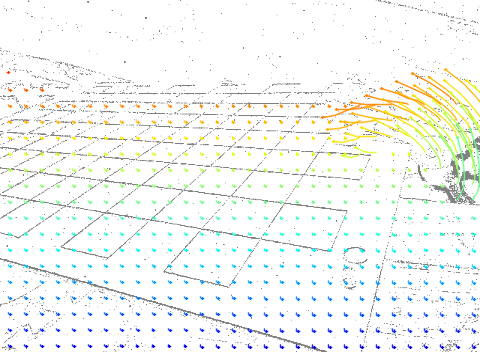}}
        &\gframe{\includegraphics[clip,trim={0cm 0cm 0cm 0cm},width=\linewidth]{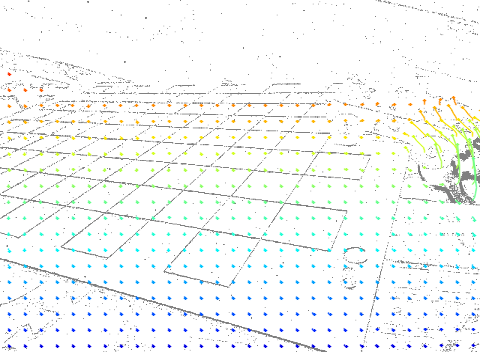}}
		&\gframe{\includegraphics[clip,trim={0cm 0cm 0cm 0cm},width=\linewidth]{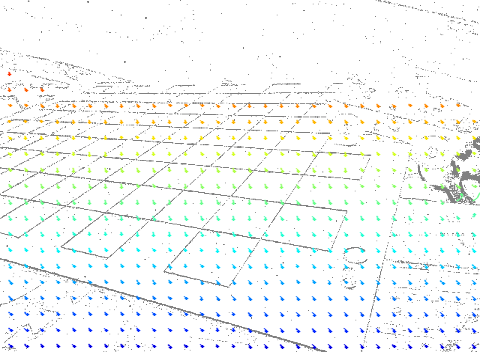}}
        &\gframe{\includegraphics[clip,trim={0cm 0cm 0cm 0cm},width=\linewidth]{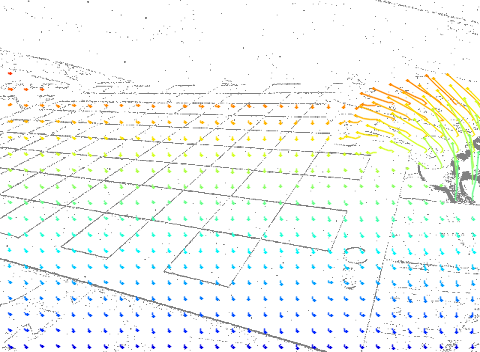}} \\

        \rotatebox{90}{\makecell{Example 9}}
		&\gframe{\includegraphics[clip,trim={0cm 0cm 0cm 0cm},width=\linewidth]{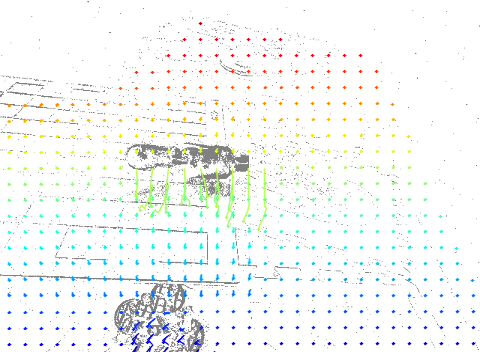}}
        &\gframe{\includegraphics[clip,trim={0cm 0cm 0cm 0cm},width=\linewidth]{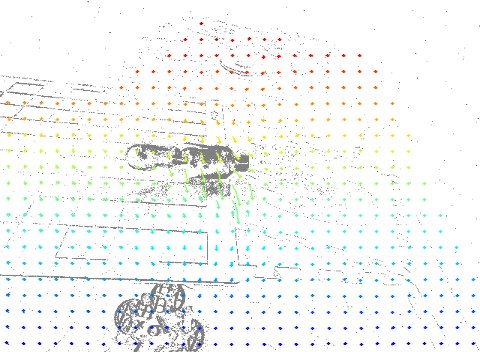}}
		&\gframe{\includegraphics[clip,trim={0cm 0cm 0cm 0cm},width=\linewidth]{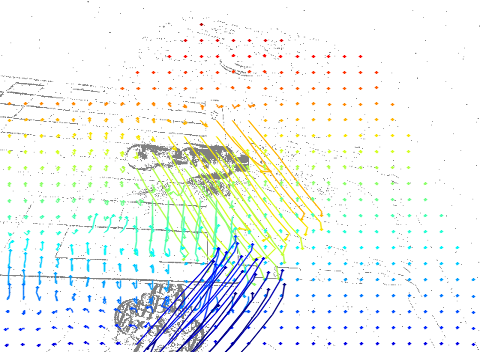}}
        &\gframe{\includegraphics[clip,trim={0cm 0cm 0cm 0cm},width=\linewidth]{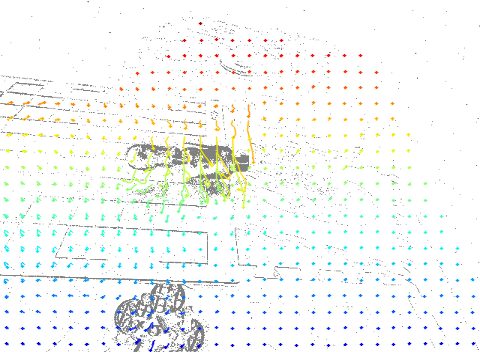}} \\

        \rotatebox{90}{\makecell{Example 10}}
		&\gframe{\includegraphics[clip,trim={0cm 0cm 0cm 0cm},width=\linewidth]{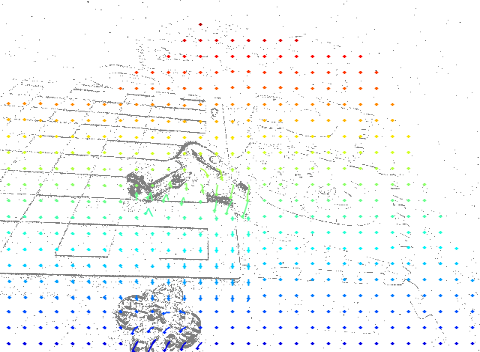}}
        &\gframe{\includegraphics[clip,trim={0cm 0cm 0cm 0cm},width=\linewidth]{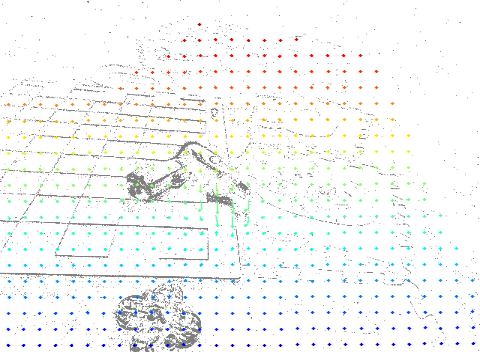}}
		&\gframe{\includegraphics[clip,trim={0cm 0cm 0cm 0cm},width=\linewidth]{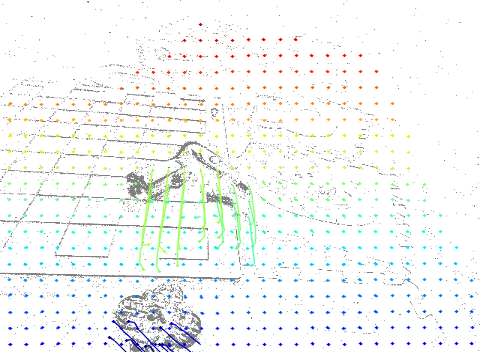}}
        &\gframe{\includegraphics[clip,trim={0cm 0cm 0cm 0cm},width=\linewidth]{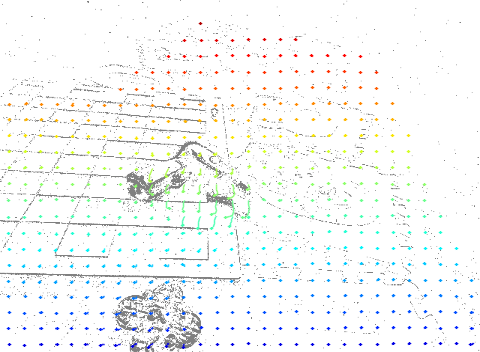}} \\

		& \textbf{(a)} GT
		& \textbf{(b)} In-domain
		& \textbf{(c)} Zero-shot
		& \textbf{(d)} Ours
	\end{tabular}
	}
	\caption{Additional results on EVIMO2. 
 Same notation as Fig.~3 in the main paper.
 \vspace{4ex}
 }
	\label{fig:supp:results_evimo2}

\def\figWidth{0.185\linewidth}
    {\scriptsize
    \setlength{\tabcolsep}{1pt}
	\begin{tabular}{
	>{\centering\arraybackslash}m{0.3cm} 
	>{\centering\arraybackslash}m{\figWidth} 
	>{\centering\arraybackslash}m{\figWidth} 
	>{\centering\arraybackslash}m{\figWidth} 
	>{\centering\arraybackslash}m{\figWidth} 
	>{\centering\arraybackslash}m{\figWidth}
    }
        \rotatebox{90}{\makecell{\tiny interlaken 00 b}}
		&\gframe{\includegraphics[clip,trim={0cm 0cm 0cm 0cm},width=\linewidth]{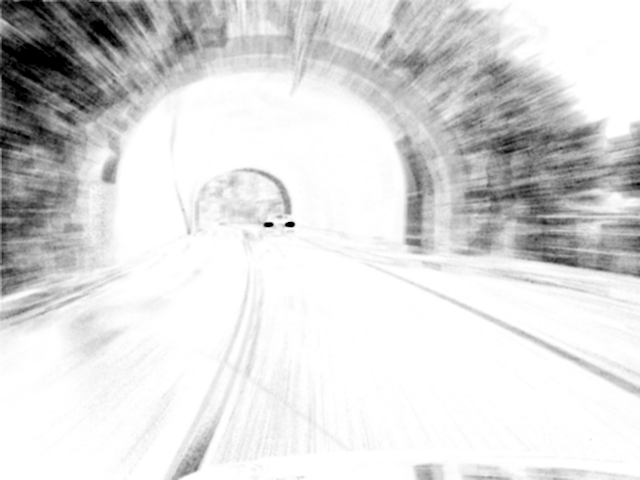}}
		&\gframe{\includegraphics[clip,trim={0cm 0cm 0cm 0cm},width=\linewidth]{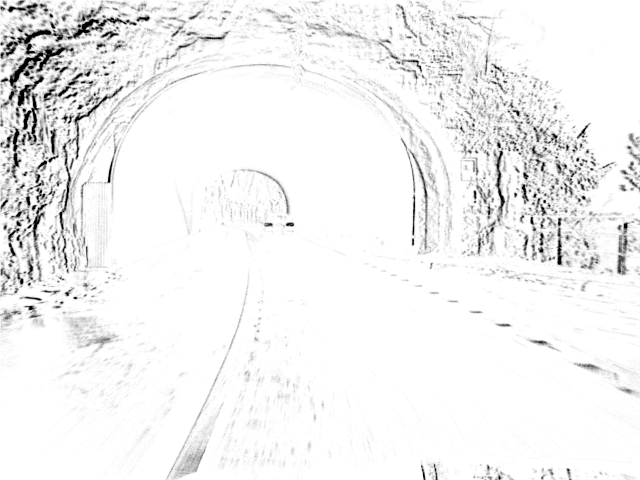}}
		&\includegraphics[clip,trim={0cm 0cm 0cm 0cm},width=\linewidth]{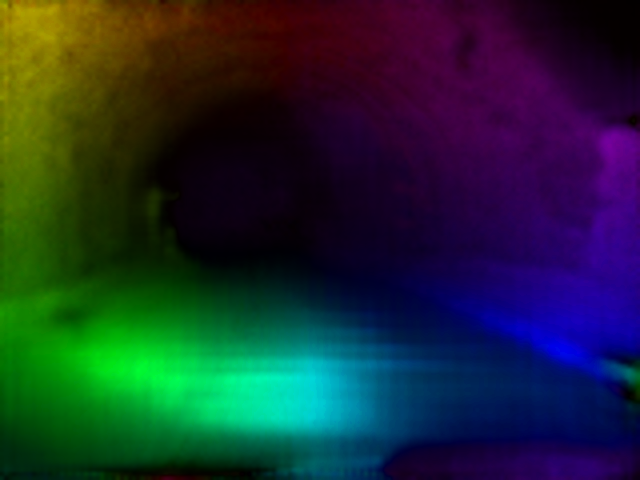}
		&\includegraphics[clip,trim={0cm 0cm 0cm 0cm},width=\linewidth]{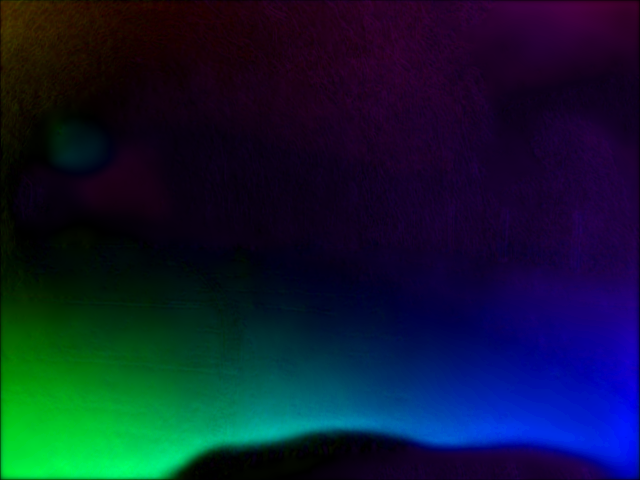}
        &\includegraphics[clip,trim={0cm 0cm 0cm 0cm},width=\linewidth]{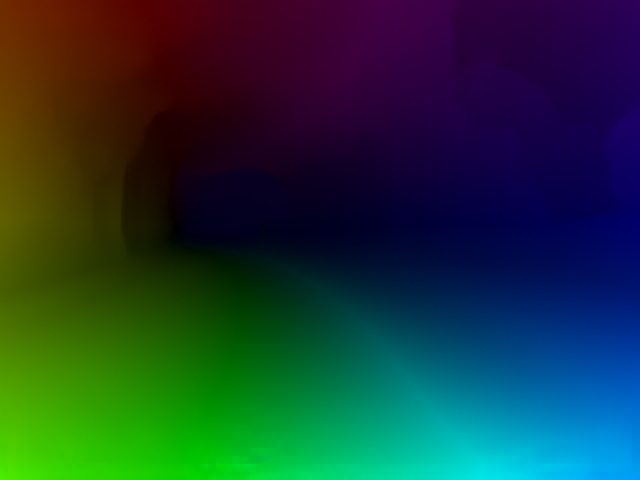} \\

        \rotatebox{90}{\makecell{\tiny thun 01 a}}
		&\gframe{\includegraphics[clip,trim={0cm 0cm 0cm 0cm},width=\linewidth]{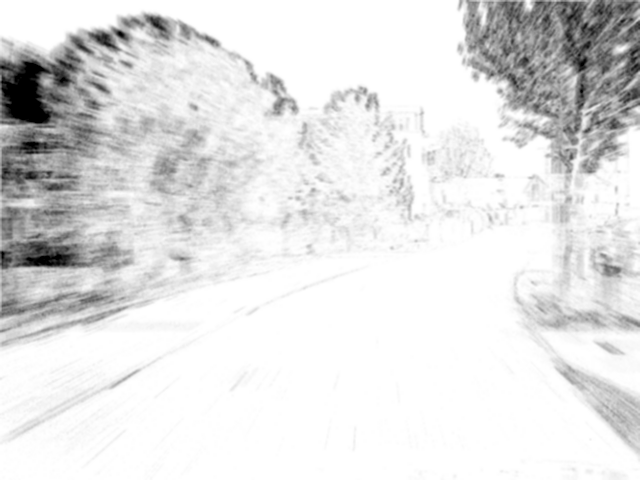}}
		&\gframe{\includegraphics[clip,trim={0cm 0cm 0cm 0cm},width=\linewidth]{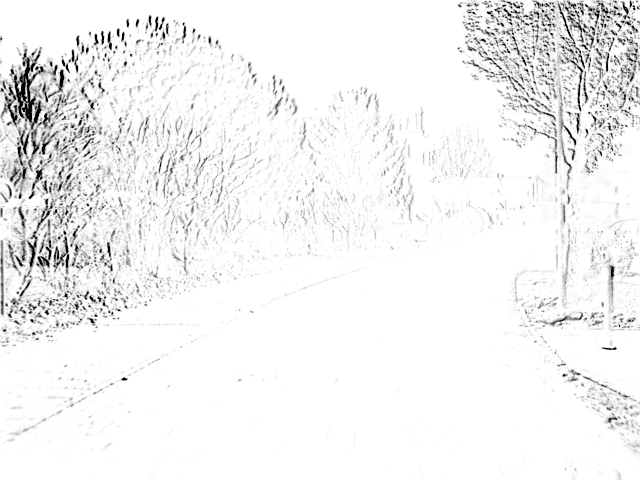}}
		&\includegraphics[clip,trim={0cm 0cm 0cm 0cm},width=\linewidth]{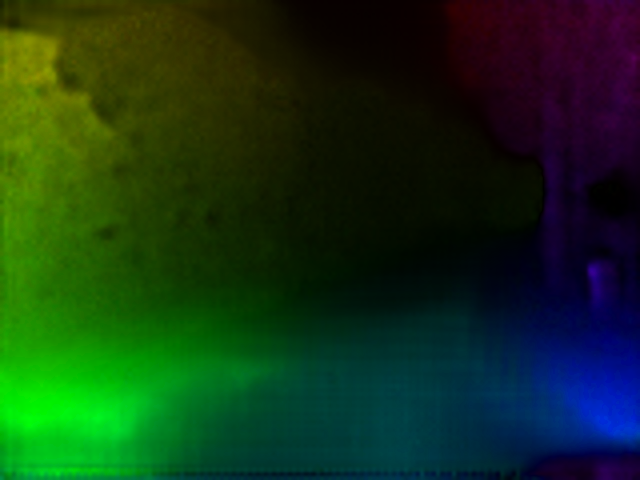}
		&\includegraphics[clip,trim={0cm 0cm 0cm 0cm},width=\linewidth]{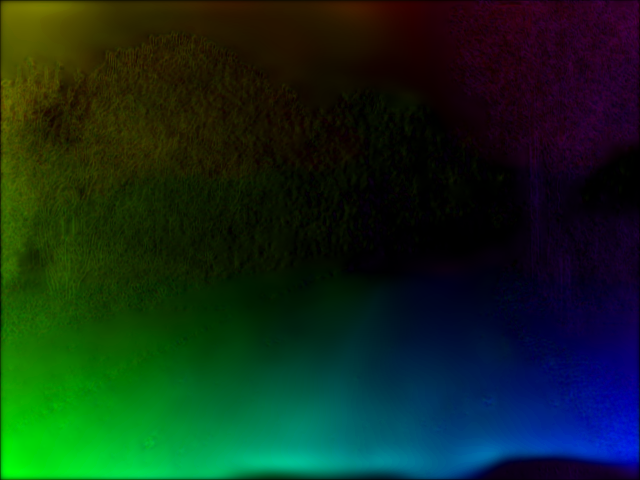}
        &\includegraphics[clip,trim={0cm 0cm 0cm 0cm},width=\linewidth]{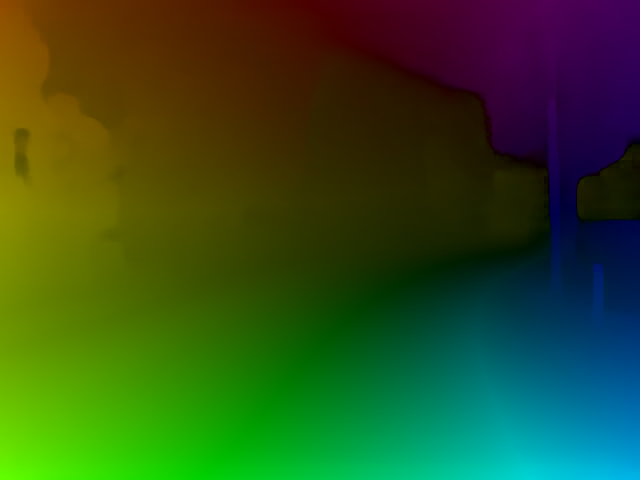} \\

        \rotatebox{90}{\makecell{\tiny thun 01 a}}
		&\gframe{\includegraphics[clip,trim={0cm 0cm 0cm 0cm},width=\linewidth]{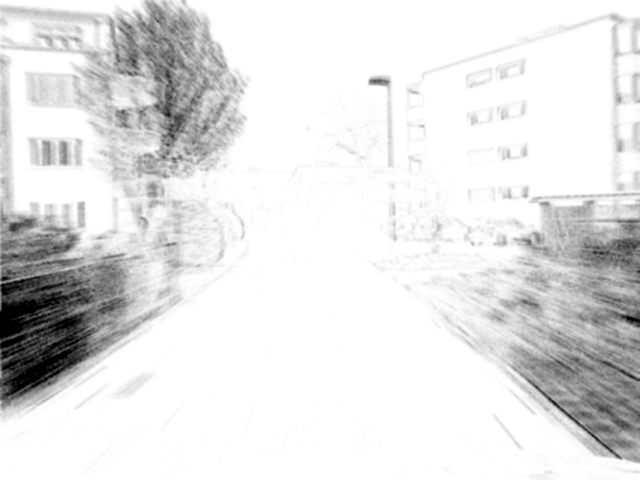}}
		&\gframe{\includegraphics[clip,trim={0cm 0cm 0cm 0cm},width=\linewidth]{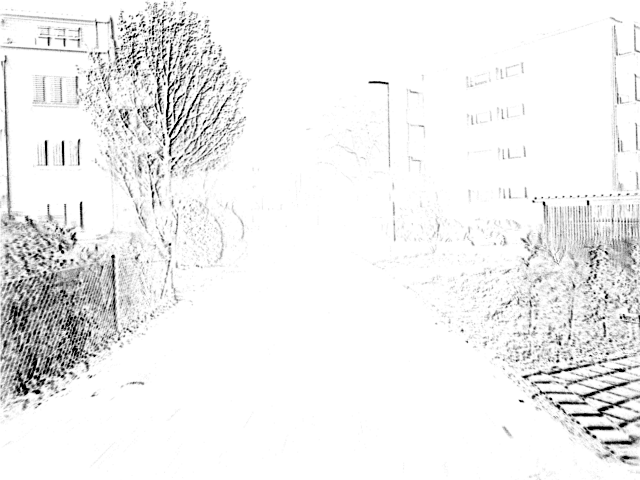}}
		&\includegraphics[clip,trim={0cm 0cm 0cm 0cm},width=\linewidth]{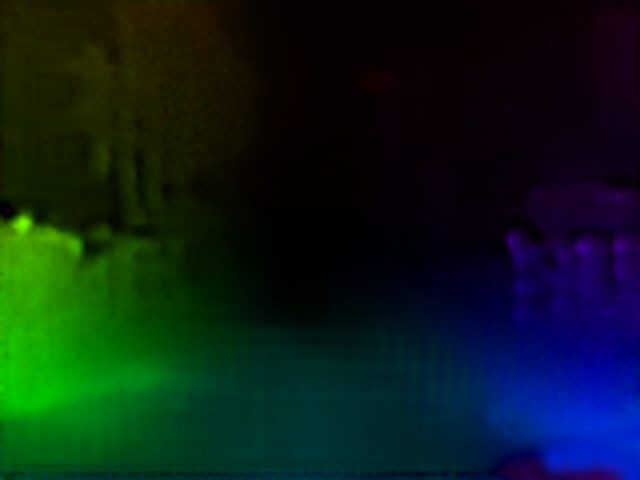}
		&\includegraphics[clip,trim={0cm 0cm 0cm 0cm},width=\linewidth]{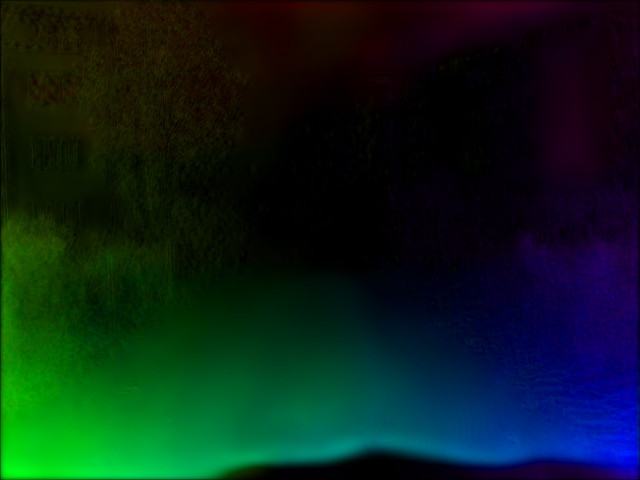}
        &\includegraphics[clip,trim={0cm 0cm 0cm 0cm},width=\linewidth]{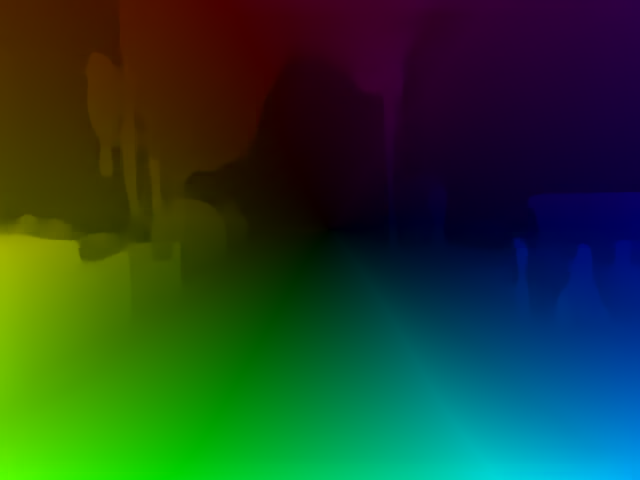} \\

        \rotatebox{90}{\makecell{\tiny zurich city 14a}}
		&\gframe{\includegraphics[clip,trim={0cm 0cm 0cm 0cm},width=\linewidth]{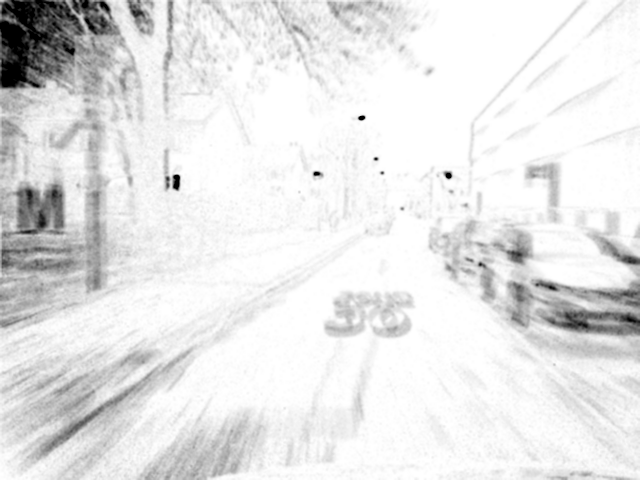}}
		&\gframe{\includegraphics[clip,trim={0cm 0cm 0cm 0cm},width=\linewidth]{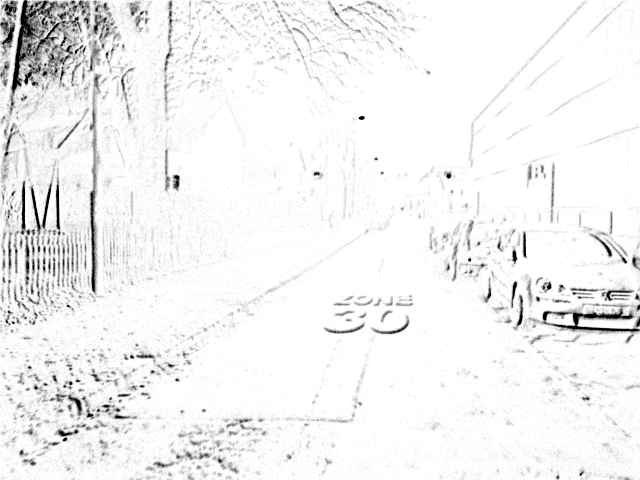}}
		&\includegraphics[clip,trim={0cm 0cm 0cm 0cm},width=\linewidth]{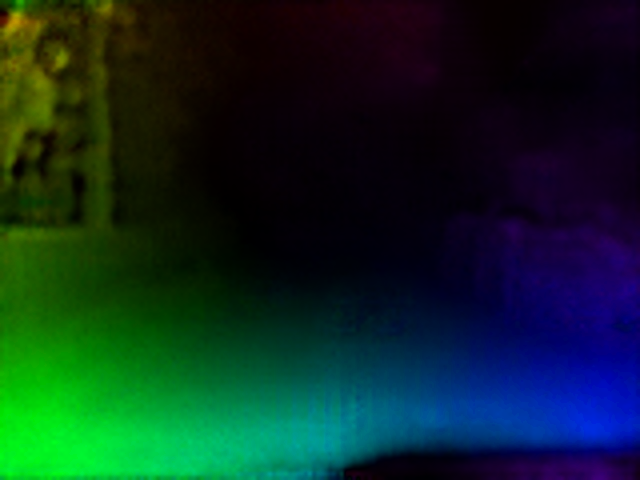}
		&\includegraphics[clip,trim={0cm 0cm 0cm 0cm},width=\linewidth]{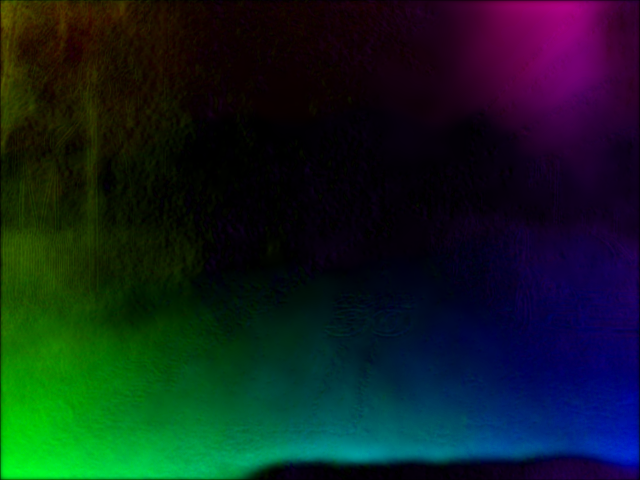}
        &\includegraphics[clip,trim={0cm 0cm 0cm 0cm},width=\linewidth]{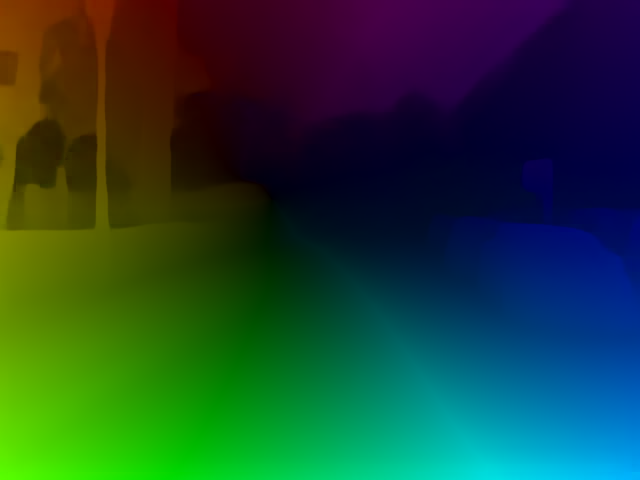} \\

		& \textbf{(a)} Events
		& \textbf{(b)} IWE (Ours)
		& \textbf{(c)} Flow (Ours)
		& \textbf{(d)} Flow Paredes~\cite{Paredes23iccv}
        & \textbf{(e)} Flow ERAFT (SL)~\cite{Gehrig21threedv}
	\end{tabular}
	}
	\caption{Additional results on DSEC. 
 Same notation as Fig.~5 in the main paper.}
    \label{fig:supp:results_dsec}
\end{figure*}


\begin{thebibliography}{10}
\providecommand{\url}[1]{\texttt{#1}}
\providecommand{\urlprefix}{URL }
\providecommand{\doi}[1]{https://doi.org/#1}

\bibitem{Akhter10pami}
Akhter, I., Sheikh, Y., Khan, S., Kanade, T.: Trajectory space: A dual
  representation for nonrigid structure from motion. {IEEE} Trans. Pattern
  Anal. Mach. Intell.  \textbf{33}(7),  1442--1456 (jul 2011).
  \doi{10.1109/TPAMI.2010.201}

\bibitem{Akolkar22pami}
Akolkar, H., Ieng, S.H., Benosman, R.: Real-time high speed motion prediction
  using fast aperture-robust event-driven visual flow. {IEEE} Trans. Pattern
  Anal. Mach. Intell.  \textbf{44}(1),  361--372 (2022).
  \doi{10.1109/TPAMI.2020.3010468}

\bibitem{Bailer17cvpr}
Bailer, C., Varanasi, K., Stricker, D.: {CNN}-based patch matching for optical
  flow with thresholded hinge embedding loss. In: {IEEE} Conf. Comput. Vis.
  Pattern Recog. (CVPR). pp. 2710--2719 (2017). \doi{10.1109/CVPR.2017.290}

\bibitem{Benosman14tnnls}
Benosman, R., Clercq, C., Lagorce, X., Ieng, S.H., Bartolozzi, C.: Event-based
  visual flow. {IEEE} Trans. Neural Netw. Learn. Syst.  \textbf{25}(2),
  407--417 (2014). \doi{10.1109/TNNLS.2013.2273537}

\bibitem{Benosman12nn}
Benosman, R., Ieng, S.H., Clercq, C., Bartolozzi, C., Srinivasan, M.:
  Asynchronous frameless event-based optical flow. Neural Netw.  \textbf{27},
  32--37 (2012). \doi{10.1016/j.neunet.2011.11.001}

\bibitem{Brebion21tits}
Brebion, V., Moreau, J., Davoine, F.: Real-time optical flow for vehicular
  perception with low- and high-resolution event cameras. {IEEE} Trans. Intell.
  Transport. Syst. pp. 1--13 (2021). \doi{10.1109/TITS.2021.3136358}

\bibitem{Brosch15fns}
Brosch, T., Tschechne, S., Neumann, H.: On event-based optical flow detection.
  Front. Neurosci.  \textbf{9}(137) (Apr 2015). \doi{10.3389/fnins.2015.00137}

\bibitem{Burner22evimo2}
Burner, L., Mitrokhin, A., Ferm\"uller, C., Aloimonos, Y.: {EVIMO2}: An event
  camera dataset for motion segmentation, optical flow, structure from motion,
  and visual inertial odometry in indoor scenes with monocular or stereo
  algorithms. ar{X}iv e-prints  (May 2022). \doi{10.48550/arXiv.2205.03467}

\bibitem{Butler12eccv}
Butler, D.J., Wulff, J., Stanley, G.B., Black, M.J.: A naturalistic open source
  movie for optical flow evaluation. In: Eur. Conf. Comput. Vis. (ECCV). pp.
  611--625 (2012). \doi{10.1007/978-3-642-33783-3_44}

\bibitem{Chaney23cvprw}
Chaney, K., Cladera~Ojeda, F., Wang, Z., Bisulco, A., Hsieh, M.A., Korpela, C.,
  Kumar, V., Taylor, C.J., Daniilidis, K.: {M3ED}: Multi-robot, multi-sensor,
  multi-environment event dataset. In: {IEEE} Conf. Comput. Vis. Pattern Recog.
  Workshops (CVPRW). pp. 4016--4023 (2023). \doi{10.1109/CVPRW59228.2023.00419}

\bibitem{Chui21arxiv}
Chui, J., Klenk, S., Cremers, D.: Event-based feature tracking in continuous
  time with sliding window optimization. In: ar{X}iv e-prints (2021).
  \doi{10.48550/arXiv.2107.04536}

\bibitem{Cuadrado23fns}
Cuadrado, J., Ran{\c{c}}on, U., Cottereau, B.R., Barranco, F., Masquelier, T.:
  Optical flow estimation from event-based cameras and spiking neural networks.
  Front. Neurosci.  \textbf{17},  1160034 (2023).
  \doi{10.3389/fnins.2023.1160034}

\bibitem{Ding22aaai}
Ding, Z., Zhao, R., Zhang, J., Gao, T., Xiong, R., Yu, Z., Huang, T.:
  Spatio-temporal recurrent networks for event-based optical flow estimation.
  In: {AAAI} Conf. Artificial Intell. vol.~36, pp. 525--533 (2022).
  \doi{10.1609/aaai.v36i1.19931}

\bibitem{Doersch22neurips}
Doersch, C., Gupta, A., Markeeva, L., Recasens, A., Smaira, L., Aytar, Y.,
  Carreira, J., Zisserman, A., Yang, Y.: Tap-vid: A benchmark for tracking any
  point in a video. Adv. Neural Inf. Process. Syst. (NeurIPS)  \textbf{35},
  13610--13626 (2022)

\bibitem{Doersch23arxiv}
Doersch, C., Yang, Y., Vecerik, M., Gokay, D., Gupta, A., Aytar, Y., Carreira,
  J., Zisserman, A.: Tapir: Tracking any point with per-frame initialization
  and temporal refinement. arXiv preprint arXiv:2306.08637  (2023)

\bibitem{Dosovitskiy15iccv}
Dosovitskiy, A., Fischer, P., Ilg, E., H{\"a}usser, P., Haz{\i}rba{\c{s}}, C.,
  Golkov, V., van~der Smagt, P., Cremers, D., Brox, T.: Flow{N}et: Learning
  optical flow with convolutional networks. In: Int. Conf. Comput. Vis. (ICCV).
  pp. 2758--2766 (2015). \doi{10.1109/ICCV.2015.316}

\bibitem{Feydy20neurips}
Feydy, J., Glaun{\`e}s, A., Charlier, B., Bronstein, M.: Fast geometric
  learning with symbolic matrices. Adv. Neural Inf. Process. Syst. (NeurIPS)
  \textbf{33},  14448--14462 (2020)

\bibitem{Finateu20isscc}
Finateu, T., Niwa, A., Matolin, D., Tsuchimoto, K., Mascheroni, A., Reynaud,
  E., Mostafalu, P., Brady, F., Chotard, L., LeGoff, F., Takahashi, H.,
  Wakabayashi, H., Oike, Y., Posch, C.: A 1280x720 back-illuminated stacked
  temporal contrast event-based vision sensor with 4.86$\mu$m pixels,
  1.066{G}eps readout, programmable event-rate controller and compressive
  data-formatting pipeline. In: {IEEE} Int. Solid-State Circuits Conf. (ISSCC).
  pp. 112--114 (2020). \doi{10.1109/ISSCC19947.2020.9063149}

\bibitem{Gallego20pami}
Gallego, G., Delbruck, T., Orchard, G., Bartolozzi, C., Taba, B., Censi, A.,
  Leutenegger, S., Davison, A., Conradt, J., Daniilidis, K., Scaramuzza, D.:
  Event-based vision: A survey. {IEEE} Trans. Pattern Anal. Mach. Intell.
  \textbf{44}(1),  154--180 (2022). \doi{10.1109/TPAMI.2020.3008413}

\bibitem{Gallego19cvpr}
Gallego, G., Gehrig, M., Scaramuzza, D.: Focus is all you need: Loss functions
  for event-based vision. In: {IEEE} Conf. Comput. Vis. Pattern Recog. (CVPR).
  pp. 12272--12281 (2019). \doi{10.1109/CVPR.2019.01256}

\bibitem{Gallego18cvpr}
Gallego, G., Rebecq, H., Scaramuzza, D.: A unifying contrast maximization
  framework for event cameras, with applications to motion, depth, and optical
  flow estimation. In: {IEEE} Conf. Comput. Vis. Pattern Recog. (CVPR). pp.
  3867--3876 (2018). \doi{10.1109/CVPR.2018.00407}

\bibitem{Gallego17ral}
Gallego, G., Scaramuzza, D.: Accurate angular velocity estimation with an event
  camera. {IEEE} Robot. Autom. Lett.  \textbf{2}(2),  632--639 (2017).
  \doi{10.1109/LRA.2016.2647639}

\bibitem{Gehrig20cvpr}
Gehrig, D., Gehrig, M., Hidalgo-Carri\'o, J., Scaramuzza, D.: {V}ideo to
  {E}vents: Recycling video datasets for event cameras. In: {IEEE} Conf.
  Comput. Vis. Pattern Recog. (CVPR). pp. 3583--3592 (2020).
  \doi{10.1109/CVPR42600.2020.00364}

\bibitem{Gehrig19iccv}
Gehrig, D., Loquercio, A., Derpanis, K.G., Scaramuzza, D.: End-to-end learning
  of representations for asynchronous event-based data. In: Int. Conf. Comput.
  Vis. (ICCV). pp. 5632--5642 (2019). \doi{10.1109/ICCV.2019.00573}

\bibitem{Gehrig21ral}
Gehrig, M., Aarents, W., Gehrig, D., Scaramuzza, D.: {DSEC}: A stereo event
  camera dataset for driving scenarios. {IEEE} Robot. Autom. Lett.
  \textbf{6}(3),  4947--4954 (2021). \doi{10.1109/LRA.2021.3068942}

\bibitem{Gehrig21threedv}
Gehrig, M., Millhäusler, M., Gehrig, D., Scaramuzza, D.: {E-RAFT}: Dense
  optical flow from event cameras. In: Int. Conf. 3D Vision (3DV). pp. 197--206
  (2021). \doi{10.1109/3DV53792.2021.00030}

\bibitem{Gehrig24pami}
Gehrig, M., Muglikar, M., Scaramuzza, D.: Dense continuous-time optical flow
  from event cameras. {IEEE} Trans. Pattern Anal. Mach. Intell. pp. 1--12
  (2024). \doi{10.1109/TPAMI.2024.3361671}

\bibitem{Guney18eccvw}
Guney, F., Sevilla-Lara, L., Sun, D., Wulff, J.: "what is optical flow for?":
  Workshop results and summary. In: Eur. Conf. Comput. Vis. Workshops (ECCVW).
  pp.~0--0 (2018)

\bibitem{Guo24tro}
Guo, S., Gallego, G.: {CMax}-{SLAM}: Event-based rotational-motion bundle
  adjustment and {SLAM} system using contrast maximization. {IEEE} Trans.
  Robot.  \textbf{40},  2442--2461 (2024). \doi{10.1109/TRO.2024.3378443}

\bibitem{Hagenaars21neurips}
Hagenaars, J., Paredes-Vall{\'e}s, F., De~Croon, G.: Self-supervised learning
  of event-based optical flow with spiking neural networks. Adv. Neural Inf.
  Process. Syst. (NeurIPS)  \textbf{34},  7167--7179 (2021)

\bibitem{Hamann22icprvaib}
Hamann, F., Gallego, G.: Stereo co-capture system for recording and tracking
  fish with frame- and event cameras. In: 26th Int. Conf. on Pattern
  Recognition (ICPR), Visual observation and analysis of Vertebrate And Insect
  Behavior (VAIB) Workshop (2022). \doi{10.48550/ARXIV.2207.07332}

\bibitem{Hamann24cvpr}
Hamann, F., Ghosh, S., Ju{\'a}rez-Mart{\'i}nez, I., Hart, T., Kacelnik, A.,
  Gallego, G.: Low-power, continuous remote behavioral localization with event
  cameras. In: {IEEE} Conf. Comput. Vis. Pattern Recog. (CVPR) (2024)

\bibitem{Harley22eccv}
Harley, A.W., Fang, Z., Fragkiadaki, K.: Particle video revisited: Tracking
  through occlusions using point trajectories. In: Eur. Conf. Comput. Vis.
  (ECCV). pp. 59--75 (2022)

\bibitem{Huang22eccv}
Huang, Z., Shi, X., Zhang, C., Wang, Q., Cheung, K.C., Qin, H., Dai, J., Li,
  H.: Flowformer: A transformer architecture for optical flow. In: Eur. Conf.
  Comput. Vis. (ECCV). pp. 668--685 (2022)

\bibitem{Ilg17cvpr}
Ilg, E., Mayer, N., Saikia, T., Keuper, M., Dosovitskiy, A., Brox, T.:
  {FlowNet} 2.0: Evolution of optical flow estimation with deep networks. In:
  {IEEE} Conf. Comput. Vis. Pattern Recog. (CVPR). pp. 1647--1655 (2017).
  \doi{10.1109/cvpr.2017.179}

\bibitem{Janai18eccv}
Janai, J., Guney, F., Ranjan, A., Black, M., Geiger, A.: Unsupervised learning
  of multi-frame optical flow with occlusions. In: Eur. Conf. Comput. Vis.
  (ECCV). pp. 690--706 (2018)

\bibitem{Jonschkowski20eccv}
Jonschkowski, R., Stone, A., Barron, J.T., Gordon, A., Konolige, K., Angelova,
  A.: What matters in unsupervised optical flow. In: Eur. Conf. Comput. Vis.
  (ECCV). pp. 557--572 (2020)

\bibitem{Karaev23arxiv}
Karaev, N., Rocco, I., Graham, B., Neverova, N., Vedaldi, A., Rupprecht, C.:
  Cotracker: It is better to track together. arXiv preprint arXiv:2307.07635
  (2023)

\bibitem{Kim21ral}
Kim, H., Kim, H.J.: Real-time rotational motion estimation with contrast
  maximization over globally aligned events. {IEEE} Robot. Autom. Lett.
  \textbf{6}(3),  6016--6023 (2021). \doi{10.1109/LRA.2021.3088793}

\bibitem{Lee20eccv}
Lee, C., Kosta, A., Zhu, A.Z., Chaney, K., Daniilidis, K., Roy, K.:
  Spike-flownet: Event-based optical flow estimation with energy-efficient
  hybrid neural networks. In: Eur. Conf. Comput. Vis. (ECCV). pp. 366--382
  (2020)

\bibitem{Lichtsteiner08ssc}
Lichtsteiner, P., Posch, C., Delbruck, T.: {A 128$\times$128 120 dB 15 $\mu$s
  latency asynchronous temporal contrast vision sensor}. {IEEE} J. Solid-State
  Circuits  \textbf{43}(2),  566--576 (2008). \doi{10.1109/JSSC.2007.914337}

\bibitem{Liu23iccv}
Liu, H., Chen, G., Qu, S., Zhang, Y., Li, Z., Knoll, A., Jiang, C.: {TMA}:
  Temporal motion aggregation for event-based optical flow. In: Int. Conf.
  Comput. Vis. (ICCV). pp. 9651--9660 (Oct 2023).
  \doi{10.1109/ICCV51070.2023.00888}

\bibitem{Liu18bmvc}
Liu, M., Delbruck, T.: Adaptive time-slice block-matching optical flow
  algorithm for dynamic vision sensors. In: British Mach. Vis. Conf. (BMVC).
  pp. 1--12 (2018)

\bibitem{Mayer16cvpr}
Mayer, N., Ilg, E., Hausser, P., Fischer, P., Cremers, D., Dosovitskiy, A.,
  Brox, T.: A large dataset to train convolutional networks for disparity,
  optical flow, and scene flow estimation. In: {IEEE} Conf. Comput. Vis.
  Pattern Recog. (CVPR). pp. 4040--4048 (2016)

\bibitem{Orchard13biocas}
Orchard, G., Benosman, R., Etienne-Cummings, R., Thakor, N.V.: A spiking neural
  network architecture for visual motion estimation. In: {IEEE} Biomed.
  Circuits Syst. Conf. ({BioCAS}). pp. 298--301 (2013).
  \doi{10.1109/biocas.2013.6679698}

\bibitem{Paredes21cvpr}
Paredes-Valles, F., de~Croon, G.C.H.E.: Back to event basics: Self-supervised
  learning of image reconstruction for event cameras via photometric constancy.
  In: {IEEE} Conf. Comput. Vis. Pattern Recog. (CVPR). pp. 3445--3454 (2021).
  \doi{10.1109/CVPR46437.2021.00345}

\bibitem{Paredes23iccv}
Paredes-Vall{\'e}s, F., Scheper, K.Y., De~Wagter, C., de~Croon, G.C.: Taming
  contrast maximization for learning sequential, low-latency, event-based
  optical flow. In: Int. Conf. Comput. Vis. (ICCV). pp. 9661--9671 (Oct 2023).
  \doi{10.1109/ICCV51070.2023.00889}

\bibitem{Paredes24scirob}
Paredes-Vallés, F., Hagenaars, J.J., Dupeyroux, J., Stroobants, S., Xu, Y.,
  de~Croon, G.C.H.E.: Fully neuromorphic vision and control for autonomous
  drone flight. Science Robotics  \textbf{9}(90),  eadi0591 (2024).
  \doi{10.1126/scirobotics.adi0591}

\bibitem{Posch14ieee}
Posch, C., Serrano-Gotarredona, T., Linares-Barranco, B., Delbruck, T.:
  Retinomorphic event-based vision sensors: Bioinspired cameras with spiking
  output. Proc. {IEEE}  \textbf{102}(10),  1470--1484 (Oct 2014).
  \doi{10.1109/jproc.2014.2346153}

\bibitem{Rebecq18corl}
Rebecq, H., Gehrig, D., Scaramuzza, D.: {ESIM}: an open event camera simulator.
  In: Conf. on Robotics Learning (CoRL). Proc. Machine Learning Research,
  vol.~87, pp. 969--982. PMLR (2018)

\bibitem{Ren17aaai}
Ren, Z., Yan, J., Ni, B., Liu, B., Yang, X., Zha, H.: Unsupervised deep
  learning for optical flow estimation. In: {AAAI} Conf. Artificial Intell.
  vol.~31 (2017)

\bibitem{Ronneberger15icmicci}
Ronneberger, O., Fischer, P., Brox, T.: {U}-{N}et: Convolutional networks for
  biomedical image segmentation. In: Int. Conf. Medical Image Computing and
  Computer-Assisted Intervention (MICCAI). pp. 234--241 (2015)

\bibitem{Sand08ijcv}
Sand, P., Teller, S.: Particle video: Long-range motion estimation using point
  trajectories. Int. J. Comput. Vis.  \textbf{80},  72--91 (2008)

\bibitem{Seok20wacv}
Seok, H., Lim, J.: Robust feature tracking in dvs event stream using {B}ezier
  mapping. In: {IEEE} Winter Conf. Appl. Comput. Vis. (WACV). pp. 1647--1656
  (2020). \doi{10.1109/WACV45572.2020.9093607}

\bibitem{Shiba22sensors}
Shiba, S., Aoki, Y., Gallego, G.: Event collapse in contrast maximization
  frameworks. Sensors  \textbf{22}(14),  1--20 (2022). \doi{10.3390/s22145190}

\bibitem{Shiba22aisy}
Shiba, S., Aoki, Y., Gallego, G.: A fast geometric regularizer to mitigate
  event collapse in the contrast maximization framework. Adv. Intell. Syst. p.
  2200251 (2022). \doi{10.1002/aisy.202200251}

\bibitem{Shiba22eccv}
Shiba, S., Aoki, Y., Gallego, G.: Secrets of event-based optical flow. In: Eur.
  Conf. Comput. Vis. (ECCV). pp. 628--645 (2022).
  \doi{10.1007/978-3-031-19797-0\_36}

\bibitem{Shiba23pami}
Shiba, S., Hamann, F., Aoki, Y., Gallego, G.: Event-based background oriented
  schlieren. {IEEE} Trans. Pattern Anal. Mach. Intell.  \textbf{46}(4),
  2011--2026 (2024). \doi{10.1109/TPAMI.2023.3328188}

\bibitem{Shiba24pami}
Shiba, S., Klose, Y., Aoki, Y., Gallego, G.: Secrets of event-based optical
  flow, depth, and ego-motion by contrast maximization. {IEEE} Trans. Pattern
  Anal. Mach. Intell. pp. 1--18 (2024). \doi{10.1109/TPAMI.2024.3396116}

\bibitem{Stoffregen20eccv}
Stoffregen, T., Scheerlinck, C., Scaramuzza, D., Drummond, T., Barnes, N.,
  Kleeman, L., Mahony, R.: Reducing the sim-to-real gap for event cameras. In:
  Eur. Conf. Comput. Vis. (ECCV). pp. 534--549 (2020).
  \doi{https://doi.org/10.1007/978-3-030-58583-9\_32}

\bibitem{Stone21cvpr}
Stone, A., Maurer, D., Ayvaci, A., Angelova, A., Jonschkowski, R.: Smurf:
  Self-teaching multi-frame unsupervised raft with full-image warping. In:
  {IEEE} Conf. Comput. Vis. Pattern Recog. (CVPR). pp. 3887--3896 (2021)

\bibitem{Sun2021cvpr}
Sun, D., Vlasic, D., Herrmann, C., Jampani, V., Krainin, M., Chang, H., Zabih,
  R., Freeman, W.T., Liu, C.: Autoflow: Learning a better training set for
  optical flow. In: {IEEE} Conf. Comput. Vis. Pattern Recog. (CVPR). pp.
  10093--10102 (2021)

\bibitem{Sun18cvpr}
Sun, D., Yang, X., Liu, M.Y., Kautz, J.: {PWC-Net}: {CNNs} for optical flow
  using pyramid, warping, and cost volume. In: {IEEE} Conf. Comput. Vis.
  Pattern Recog. (CVPR) (2018)

\bibitem{Sun24arxiv}
Sun, X., Harley, A.W., Guibas, L.J.: Refining pre-trained motion models. arXiv
  preprint arXiv:2401.00850  (2024)

\bibitem{Teed20eccv}
Teed, Z., Deng, J.: {RAFT}: Recurrent all pairs field transforms for optical
  flow. In: Eur. Conf. Comput. Vis. (ECCV). pp. 402--419 (2020).
  \doi{10.1007/978-3-030-58536-5\_24}

\bibitem{Tulyakov22cvpr}
Tulyakov, S., Bochicchio, A., Gehrig, D., Georgoulis, S., Li, Y., Scaramuzza,
  D.: Time lens++: Event-based frame interpolation with parametric non-linear
  flow and multi-scale fusion. In: {IEEE} Conf. Comput. Vis. Pattern Recog.
  (CVPR). pp. 17755--17764 (Jun 2022)

\bibitem{Valmadre12cvpr}
Valmadre, J., Lucey, S.: General trajectory prior for non-rigid reconstruction.
  In: {IEEE} Conf. Comput. Vis. Pattern Recog. (CVPR). pp. 1394--1401 (2012)

\bibitem{Wan22tip}
Wan, Z., Dai, Y., Mao, Y.: Learning dense and continuous optical flow from an
  event camera. {IEEE} Trans. Image Process.  \textbf{31},  7237--7251 (2022)

\bibitem{Wang21arxiv}
Wang, C., Eckart, B., Lucey, S., Gallo, O.: Neural trajectory fields for
  dynamic novel view synthesis. arXiv preprint arXiv:2105.05994  (2021)

\bibitem{Wang23arxiv_cc}
Wang, Z., Hamann, F., Chaney, K., Jiang, W., Gallego, G., Daniilidis, K.:
  Event-based continuous color video decompression from single frames. arXiv
  preprint arXiv:2312.00113  (2023)

\bibitem{Weikersdorfer13icvs}
Weikersdorfer, D., Hoffmann, R., Conradt, J.: Simultaneous localization and
  mapping for event-based vision systems. In: Int. Conf. Comput. Vis. Syst.
  (ICVS). pp. 133--142 (2013). \doi{10.1007/978-3-642-39402-7_14}

\bibitem{Wu22arxiv}
Wu, Y., Paredes-Vall{\'e}s, F., de~Croon, G.C.: Lightweight event-based optical
  flow estimation via iterative deblurring. arXiv preprint arXiv:2211.13726
  (2022)

\bibitem{Ye19arxiv}
Ye, C., Mitrokhin, A., Parameshwara, C., Ferm\"uller, C., Yorke, J.A.,
  Aloimonos, Y.: Unsupervised learning of dense optical flow, depth and
  egomotion with event-based sensors. In: IEEE/RSJ Int. Conf. Intell. Robot.
  Syst. (IROS). pp. 5831--5838 (2020). \doi{10.1109/IROS45743.2020.9341224}

\bibitem{young2014image}
Young, P., Lai, A., Hodosh, M., Hockenmaier, J.: From image descriptions to
  visual denotations: New similarity metrics for semantic inference over event
  descriptions. Trans. Assoc. Computational Linguistics  \textbf{2},  67--78
  (2014)

\bibitem{Yu16eccvw}
Yu, J.J., Harley, A.W., Derpanis, K.G.: Back to basics: Unsupervised learning
  of optical flow via brightness constancy and motion smoothness. In: Eur.
  Conf. Comput. Vis. Workshops (ECCVW). pp. 3--10 (2016)

\bibitem{Zen12eccv}
Zen, G., Ricci, E., Sebe, N.: Exploiting sparse representations for robust
  analysis of noisy complex video scenes. In: Eur. Conf. Comput. Vis. (ECCV).
  pp. 199--213 (2012)

\bibitem{Zheng23cvpr}
Zheng, Y., Harley, A.W., Shen, B., Wetzstein, G., Guibas, L.J.: {PointOdyssey}:
  A large-scale synthetic dataset for long-term point tracking. In: Int. Conf.
  Comput. Vis. (ICCV). pp. 19855--19865 (2023)

\bibitem{Zhu18ral}
Zhu, A.Z., Thakur, D., Ozaslan, T., Pfrommer, B., Kumar, V., Daniilidis, K.:
  The multivehicle stereo event camera dataset: An event camera dataset for
  {3D} perception. {IEEE} Robot. Autom. Lett.  \textbf{3}(3),  2032--2039 (Jul
  2018). \doi{10.1109/lra.2018.2800793}

\bibitem{Zhu18rss}
Zhu, A.Z., Yuan, L., Chaney, K., Daniilidis, K.: {EV-FlowNet}: Self-supervised
  optical flow estimation for event-based cameras. In: Robotics: Science and
  Systems (RSS). pp.~1--9 (2018). \doi{10.15607/RSS.2018.XIV.062}

\bibitem{Zhu19cvpr}
Zhu, A.Z., Yuan, L., Chaney, K., Daniilidis, K.: Unsupervised event-based
  learning of optical flow, depth, and egomotion. In: {IEEE} Conf. Comput. Vis.
  Pattern Recog. (CVPR). pp. 989--997 (2019). \doi{10.1109/CVPR.2019.00108}

\bibitem{Zhu13pami}
Zhu, Y., Lucey, S.: Convolutional sparse coding for trajectory reconstruction.
  {IEEE} Trans. Pattern Anal. Mach. Intell.  \textbf{37}(3),  529--540 (2013)

\end{thebibliography}
\end{document}